\pdfoutput=1

\documentclass[11pt]{article}

\usepackage[]{acl}

\usepackage{times}
\usepackage{latexsym}

\usepackage[T1]{fontenc}

\usepackage[utf8]{inputenc}

\usepackage{microtype}

\usepackage{inconsolata}

\usepackage{pifont}

\usepackage{colortbl}

\usepackage{amsmath}
\usepackage{amsfonts}
\usepackage{amssymb}

\usepackage{changepage}

\usepackage{microtype}
\usepackage{adjustbox}
\usepackage{booktabs}
\usepackage{bbm}

\usepackage{multirow}
\usepackage{graphicx}

\usepackage{gb4e}
\noautomath

\usepackage{enumitem}

\usepackage[normalem]{ulem}

\newcommand{\namedref}[2]{\hyperref[#2]{#1~\ref*{#2}}}
\newcommand{\sectionref}[1]{\namedref{Section}{#1}}
\newcommand{\tableref}[1]{\namedref{Table}{#1}}
\newcommand{\figureref}[1]{\namedref{Figure}{#1}}
\newcommand{\appendixref}[1]{\namedref{Appendix}{#1}}
\newcommand{\exampleref}[1]{(\ref{#1})}

\newcommand{\supported}{\texttt{supported}}
\newcommand{\contradicted}{\texttt{contradicted}}
\newcommand{\unsupported}{\texttt{unsupported}}
\newcommand{\inconclusive}{\texttt{inconclusive}}

\newcommand{\factscore}{\textsc{FActScore}}
\newcommand{\fsbio}{Biography}
\newcommand{\name}{\textsc{VeriScore}}
\newcommand{\SAFE}{\textsc{Safe}}

\newcommand{\lightgreen}[1]{\colorbox{green!20}{#1}}
\newcommand{\lightred}[1]{\colorbox{red!20}{#1}}

\usepackage[]{listings}
\title{\name: Evaluating the factuality of \\verifiable claims in long-form text generation}

\author{Yixiao Song$^\diamondsuit$$^\spadesuit$ \quad Yekyung Kim$^\diamondsuit$ \quad  Mohit Iyyer$^\diamondsuit$ \\
$^\diamondsuit$ Manning College of Information and Computer Sciences, UMass Amherst\\
$^\spadesuit$ Department of Linguistics, UMass Amherst\\
\texttt{\{yixiaosong, yekyungkim, miyyer\}@umass.edu}
}

\begin{document}
\maketitle

\begin{abstract}

Existing metrics for evaluating the factuality of long-form text, such as \factscore~\citep{factscore} and \SAFE~\citep{longfact}, decompose an input text into ``atomic claims'' and verify each against a knowledge base like Wikipedia. These metrics are not suitable for most generation tasks because they assume that every claim is \emph{verifiable} (i.e., can plausibly be proven true or false). We address this issue with \name,\footnote{Code and data are available at \url{https://github.com/Yixiao-Song/VeriScore}.} a metric for diverse long-form generation tasks that contain both verifiable and unverifiable content. \name\ can be effectively implemented with either closed or fine-tuned open-weight language models, and human evaluation confirms that \name's extracted claims are more sensible than those from competing methods across eight different long-form tasks.  We use \name\ to evaluate generations from 16 different models across multiple long-form tasks and find that while GPT-4o is the best-performing model overall, open-weight models such as Mixtral-$8\times22$ are closing the gap. 
We show that an LM's \name\ on one task (e.g., biography generation) does not necessarily correlate to its \name\ on a different task (e.g., long-form QA), highlighting the need for expanding factuality evaluation across tasks with varying fact density.

 
\end{abstract}

\section{Introduction}

Modern approaches for evaluating the factuality of LLM-generated long-form text, such as \factscore~\citep{factscore} and \SAFE~\citep{longfact}, proceed in three stages: (1) \emph{decomposition} of the text into a list of ``atomic'' (i.e., short) claims; (2) \emph{retrieval} of relevant evidence for each claim from Wikipedia or Google Search; and (3) \emph{verification} of each claim against the retrieved evidence. These approaches implicitly assume that the input text can be entirely decomposed into \emph{atomic} and \emph{verifiable} claims. 

Unfortunately, these assumptions do not always apply to complex generation tasks such as long-form question answering (LFQA) for two reasons. First, outputs for the biography generation task studied in \factscore\ rarely go beyond introducing entities and events. However, in tasks like LFQA, we observe more complex assertions that cannot be made  ``atomic'' without losing critical context, as in:

\begin{exe}
\fontsize{10}{12}\selectfont
\vspace{-3pt}
\ex\label{ex.AJohnson}
    The impeachment of Andrew Johnson set a precedent that impeachment should be reserved for clear cases of serious misconduct rather than political disagreements. 
\vspace{-3pt}
\end{exe}

\begin{figure*}[ht!]
    \centering
    \includegraphics[width=\linewidth]{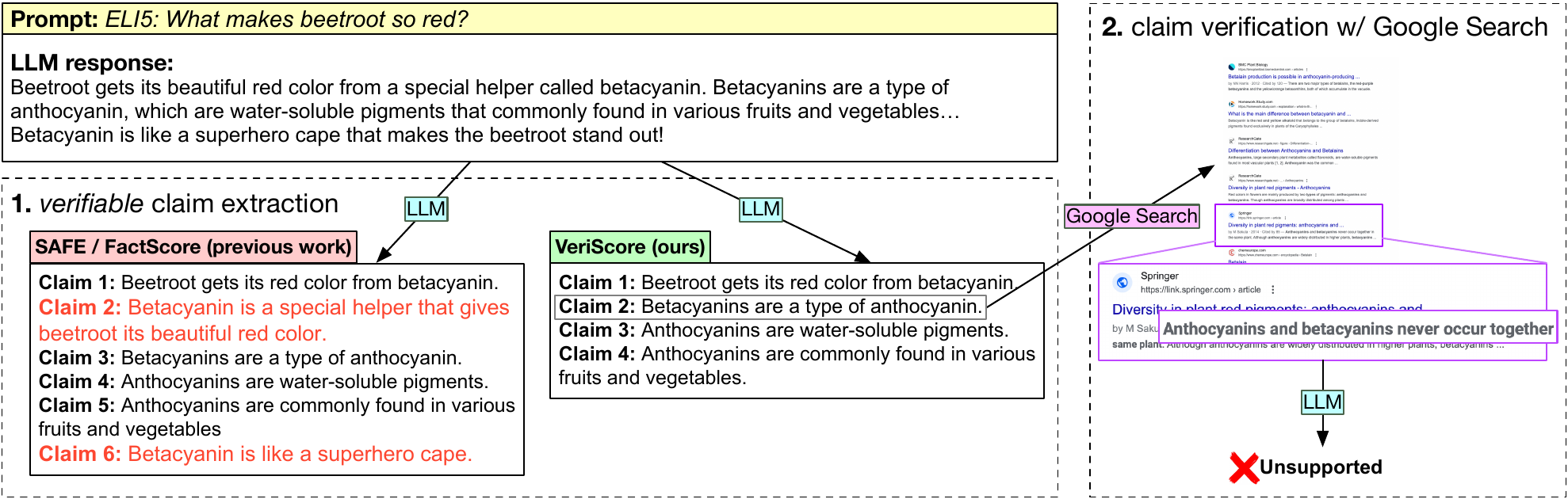}
    \caption{The pipeline of \name\, involving claim extraction and claim verification with Google Search. 
    \name\ extracts \textit{verifiable} claims. Each claim is used as a search query to retrieve evidence via Google Search, and an LLM then verifies the claim against the search results. We also show SAFE's extracted claims from the same text to highlight its propensity to extract unverifiable claims (Claim 2 and 6); see \sectionref{subsubsec:issues_in_safe} for more discussion.}
    \label{fig:veriscore}
    \vspace{-7pt}
\end{figure*}

\noindent Second, many long-form outputs interleave factual claims with unverifiable content, such as \emph{Betacyanin is like a superhero cape} in \figureref{fig:veriscore}. Since \factscore~and \SAFE~ assume all claims are verifiable, they extract everything from the text (including unverifiable content like examples or hypotheticals), which can unfairly penalize models during the final aggregation process. As such, these metrics are limited  to fact-dense and formulaic text (e.g., biographies). 


We address these issues by developing \name, an automatic metric that assesses models' factuality against Google Search results. \name's decomposition and verification steps are initially implemented using few-shot prompting, and extensive human studies confirm the quality of both steps. Subsequently, open-weight LLMs are fine-tuned on data generated by GPT-4\footnote{Henceforth, unless otherwise specified, GPT-4 refers to \texttt{gpt-4-0125-preview} and Claude 3 refers to \texttt{claude-3-opus-20240229}.} and GPT-4o to create a cost-efficient and reliable implementation. Compared to \factscore~and \SAFE, \name\ introduces two key improvements. First, \name\ only extracts what we term \textit{verifiable claims}, unlike \factscore~and \SAFE~which decompose \textit{everything}. Second, \name\ is the first approach that considers inter-sentence context when extracting claims, removing the need for expensive claim revision steps present in \SAFE\ (see \sectionref{subsubsec:issues_in_safe}). In a human study, our extraction method is preferred 93\% of the time over \SAFE's, even in biography generation.



To benchmark \name, we gather fact-seeking prompts from eight diverse domains that require long-form responses, ranging from the fact-dense biography generation task of \factscore\ to the multi-task, open-domain ShareGPT dataset for instruction-following. 
We evaluate sixteen closed and open-weight LMs with \name\ and find that GPT-4o generates the most factually-supported text when averaged across all datasets. Our analyses highlight that (1) multiple tasks (not just biography generation) are needed for comprehensive long-form factuality evaluation because an LLM's factuality varies depending on the task and domain;
and (2) verifying complex, lengthy assertions (common in many long-form tasks such as LFQA) against Google Search results can fail due to challenges in retrieving relevant documents from such queries. 

\section{\name: an automatic factuality metric}

This section details the \name\ pipeline, covering claim extraction, evidence retrieval, claim verification, and score calculation.

\subsection{Claim extraction}\label{subsec:define_claim_extraction}

Claim extraction facilitates factuality evaluation by decomposing sentences with potentially multiple independent facts \citep{factscore,tang2024minicheck}. We first examine the shortcomings of \factscore~and \SAFE~before developing a new method that focuses on extracting \textit{verifiable} claims. 


\subsubsection{Issues with claim extraction in \factscore~and SAFE}\label{subsubsec:issues_in_safe}

\factscore~\citep{factscore} extracts \textit{atomic facts}---``short statements that each contains one piece of information''.
However, their extraction method is optimized for biographies and is inapplicable to other domains. First, it does not resolve pronouns: for example, it extracts ``His notable film credits include The Game.'' from an LLM-generated biography of Lanny Flaherty.  Second, it extracts \textit{everything} instead of just verifiable claims, an issue that is inherited by \SAFE~as in \figureref{fig:veriscore}.

 
\SAFE~\citep{longfact} targets domains beyond biography and adapts \factscore's extraction prompt for a three-step pipeline: (1) claim extraction, (2) claim revision to resolve vague references, and (3) a relevance check to decide if a claim is worth checking. While~\citet{longfact} proclaim \SAFE's superior performance, a  closer inspection reveals four issues. First, besides adding a brief task description, \SAFE uses \factscore's prompt without changes. Second, the revision and relevance check adds significant processing time and cost.\footnote{\label{fn_time_consuming}Processing 100 claims without parallelization takes 35 minutes using GPT-4. SAFE's prompt templates of claim revision and relevance check alone, without filling in content, cost about \$1.7 per 100 claims (estimated using \url{https://platform.openai.com/tokenizer}).} Third, the relevance check unexpectedly removes verifiable claims. Lastly, \SAFE's generalizability is questionable given that it is only evaluated on \factscore's biography data. More details of these issues are in \appendixref{appendix:issue_in_safe}.

\subsubsection{\name's extraction approach}\label{subsubsec:our_extractor}

\factscore\ and \SAFE~ extract atomic claims with the implicit assumption that all claims are verifiable; unfortunately, this leads to the extraction of unverifiable claims (e.g., Claim 2 and 6 in \figureref{fig:veriscore}). Achieving atomicity is also hard as exemplified by Example \exampleref{ex.AJohnson}. Hence, we instead aim to extract only \textit{verifiable} claims. Inspired by frameworks of \textit{events} and \textit{states} in linguistics \citep{Maienborn, EandS}, we use the following description as a guideline: 

\begin{adjustwidth}{0.25cm}{0.25cm}
\noindent \underline{Verifiable claims} describe a single \textit{event} or \textit{state}\footnote{Event: change of state, for example, ``Jensen Huang founded NVIDIA in 1993 in California, U.S.'' State: for example, ``Westborough is a town in Worcester, MA.''} with all necessary modifiers (e.g., spatial, temporal, or relative clauses) that help denote entities or events in the real world.
\end{adjustwidth}


\noindent Formally, our claim extraction process produces a set of verifiable claims $C = \{c_1, c_2, ..., c_n\}$ from a model response $r$ where $c$ consists of meaningful parts $p$ such that $c = \{p_1, p_2, ..., p_n\}$. Each $p$ does not have to be a full proposition.\footnote{For example, if $c =$ ``Jensen Huang founded NVIDIA in 1993 in California, U.S.'', $p_1=$ \textit{Jensen Huang founded NVIDIA}, $p_2 = $ \textit{in 1993}, and $p_3 = $ \textit{in California, U.S.}}

To address the issues in \factscore~and SAFE, we design few-shot claim extraction prompts given in \appendixref{appendix:our_claim_extraction_prompts}. A sliding window, formatted as \texttt{(context1: 0-3 sentences) <SOS>focused sentence<EOS> (context2: 0-1 sentence)} is used to guide LLMs to extract verifiable claims from the focused sentence, using the context to ensure the claims are self-contained (e.g., pronouns are resolved).\footnote{For QA tasks, we always prepend the question to the sliding window. For non-QA tasks, we prepend the first sentence of a paragraph to the sliding window if the paragraph is longer than five sentences to mitigate lack-of-context issues.} Unverifiable content such as advice, fictional stories, or subjective opinions are ignored.

A human evaluation study detailed in \sectionref{subsec:1stHumanEval} confirms the advantages of our extraction method. It effectively addresses the issue of unresolved referents and eliminates the need of claim revision and removal. Additionally, our method correctly avoids extracting claims from non-factual content. 

\subsection{Evidence retrieval}\label{subsec:define_evidence_retrieval}

As in \SAFE, we use Google Search via the Serper API\footnote{\url{https://serper.dev/}} to retrieve evidence. For a claim $c\in C$, we use $c$ as the search query and retrieve the top $n$ search results $E_c = \{e_1, e_2, ..., e_n\}$ $(n \leq 10)$. We use the title, snippet, and the link of each search result returned by Serper and combine the results into an evidence list as in \citet{vu2023freshllms}.

\subsection{Claim verification}\label{subsec:define_claim_verification}

Claim verification judges whether a claim $c$ is \supported\ or  \contradicted\ by a corresponding evidence list $E_c$, or alternatively whether the verification is \inconclusive. For a claim to be \supported, all parts of the claim need to be supported (i.e., no evidence $e \in E_c$ can contradict any $p \in c$). For a claim to be contradicted, at least one $p \in c$ is contradicted by some  evidence $e \in E_c$.
\texttt{Inconclusive} cases can be classified into two types: (1) at least one part of the claim is neither supported nor contradicted with respect to $E_c$; or (2) at least one part of the claim is both supported and contradicted by different evidences $e \in E_c$. A formal definition of the three scenarios is given in \tableref{tab:claim_veri_def}. In \sectionref{sec:claim.verification}'s human study, we notice that there are very few claims directly contradicted by the evidence list $E_c$. Hence, for all subsequent experiments we combine \contradicted~and \inconclusive~into a single \unsupported~category, which renders claim verification as a binary classification task. 

\subsection{Score calculation}\label{subsec:define_score_aggregation}

An ideal generation should have both high factual precision (i.e., low hallucination) and high factual recall (i.e., not be too short or incomplete). We adopt the $F_1@K$ metric from \SAFE, which considers both factual precision and recall. $K$ is the minimum number of factual claims a model response must contain to achieve perfect recall. For each tested domain, we set $K$ as the median number of extracted facts among all model responses.

Let $\mathcal{M}$ be a language model to be evaluated and $\mathcal{X}$ be a set of prompts of a given domain. Let $r = \mathcal{M}_x$ be a response of $\mathcal{M}$ to $x \in X$, and let the transitive predicate \texttt{support}$(a,b)$ take a value of either 1 or 0. $S(r) = \frac{1}{|C|} \sum_{c\in C} \texttt{support}(c, E_c)$ is the number of supported claims of $r$. $P(r) = \frac{S(r)}{|C|}$ and $R(r) = min(\frac{S(r)}{K}, 1)$ are precision and recall. \name\ of $\mathcal{M}$ is the average of the responses' $F1@K$  within each domain, defined as: 



\fontsize{8}{10}\selectfont

\begin{equation*}
  F_1@K(r) =
      \begin{cases}
      \frac{2P(r)R_K(r)}{P(r) + R_K(r)} & \text{if } S(r) > 0\\
      0 & \text{if } S(r) = 0
    \end{cases}
\end{equation*}
\vspace{-1.8em}

\[\text{\name} = \frac{1}{|X|} \sum_{x\in X} F_1@K(\mathcal{M}_x) \]

\normalsize

\section{Validation of \name's claim extraction and verification}\label{subsec:claim.extraction}

\begin{table*}[t]
\fontsize{8}{10}\selectfont
\resizebox{\textwidth}{!}{%
\begin{tabular}{@{}p{2.5cm}p{10cm}p{0.7cm}p{0.7cm}p{3cm}@{}}
\toprule
\textbf{Name}    & \textbf{Description}   & \textbf{Usage} & \textbf{VerRatio} & \textbf{Source} \\ \midrule
Scruples       & Community judgements on real-life anecdotes from \href{https://reddit.com/r/AmItheAsshole}{r/AmItheAsshole} from November 2018 to April 2019 & HE & ---  & \citet{lourie2021scruples} \\[1mm]

CommonCrawl    & A corpus of raw web page data, metadata extracts, and text extracts & HE & --- & \href{https://commoncrawl.org/overview}{CommonCrawl} \\[1mm]

wikitext-103   & Wikipedia articles & HE & --- & \citet{merity2016pointer}   \\[1mm]

WritingPrompts [WP] & Story premises and stories written by online users on \href{https://reddit.com/r/WritingPrompts/}{r/WritingPrompts} & HE/Dev & 0.03 &  \citet{fan-etal-2018-hierarchical} \\[1mm]

\href{https://huggingface.co/datasets/anon8231489123/ShareGPT_Vicuna_unfiltered}{ShareGPT} [S.GPT]       & User-shared conversations (prompts and responses) with ChatGPT on \url{ShareGPT.com} & HE/Dev & 0.92 & \citet{chiang2023vicuna}     \\[1mm]

ELI5           & Questions and layperson-friendly answers posted on \href{https://www.reddit.com/r/explainlikeimfive/}{r/explainlikeimfive} & HE/Dev & 1.71 & Scraped by \citet{xu-etal-2023-critical}      \\[1mm]

AskHistorians [AskH]  &  Questions and answers on history topics posted on \href{https://www.reddit.com/r/AskHistorians/}{r/AskHistorians}   & HE/Dev & 1.90 & Same as above \\[1mm]

\fsbio  [Bio]    & Biography text generated by PerplexityAI, InstructGPT, and ChatGPT & HE/Dev & 2.08 &\citet{factscore}\\

LongFact  [LF]    & A prompt set of 38 topics generated by GPT-4; each topic has prompts about object \& concept; we randomly sampled 5 object and 5 concept prompts from 10 topics & Dev & 2.24 & \citet{longfact} \\[2mm]

FreshQA        & A dynamic QA benchmark whose answers can change w.r.t. updated world knowledge; we randomly sampled 200 questions with a true premise from the never- and fast-changing categories in the test set of the April 1\textsuperscript{st} version  & Dev & 1.00 & \citet{vu2023freshllms} \\[2mm]

FreshBooks  [FBs]    & We collected 20 non-fictional books that are published in 2023 and 2024. Ten paragraphs are taken from each book. LLMs are prompted to generate a continuation given a paragraph & Dev & 2.31 & Current paper;\newline Details in \tableref{tab:new_book_dataset} \\\bottomrule
\end{tabular}%
}
\caption{Datasets used in the human evaluation of claim extraction (Usage = HE) in \sectionref{subsec:1stHumanEval} and in the \name\ development (Usage = Dev) in \sectionref{subsec:claim.extraction} and \ref{subsec:fine-tuned_models}. Short name of each dataset is in square brackets. VerRatio column presents the ratio of verifiable claims to sentences per domain of GPT-4 generated responses. 
}
\label{tab:seven_datasets_for_claim_extraction}
\vspace{-7pt}
\end{table*}

\begin{figure}[t]
    \centering
    \includegraphics[scale=0.43]{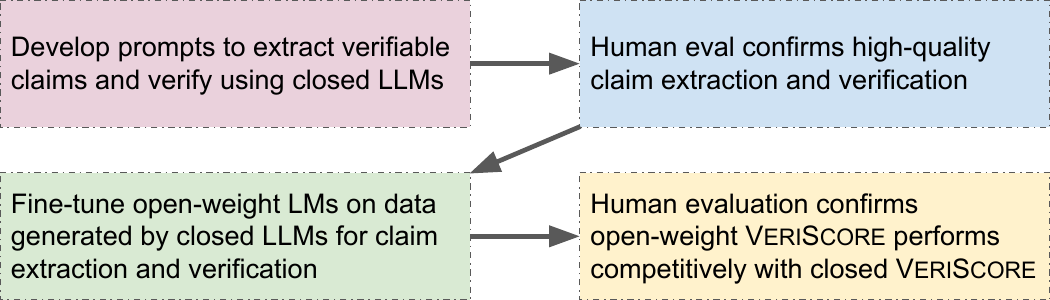}
    \caption{The development of the open-weight \name. Details in \sectionref{subsec:claim.extraction} and \sectionref{sec:veriscore_16_models_6_domains}.}\label{fig:sec3_dev_pipeline}
    \vspace{-9pt}
\end{figure}

\SAFE~and  \factscore~use closed LLMs for claim extraction and verification. Following them, as shown in \figureref{fig:sec3_dev_pipeline}, we develop \name's extraction and verification in \sectionref{subsec:claim.extraction} by prompting closed LLMs, whose effectiveness is verified by human evaluations. To mitigate the high cost of closed LLMs, we develop a free alternative in \sectionref{subsec:fine-tuned_models} by fine-tuning open-weight LLMs on data from GPT-4 and GPT-4o.

\subsection{Human evaluation on claim extraction}\label{subsec:1stHumanEval}

To verify our extraction method's efficacy, we conducted a pairwise comparison of claims extracted by \name\ and \SAFE~with three human raters. They were asked to choose the claim lists with least unverifiable content. Half of the claims were extracted by GPT-4 and the other half by Claude 3. Our method outperforms \SAFE~regardless of the model used. Because GPT-4 was preferred more often than Claude 3 with our prompts, we use GPT-4 as the claim extractor in \sectionref{sec:veriscore_16_models_6_domains}.
\vspace{-5pt}
\paragraph{Setup:} We extracted claims from 15 randomly sampled long-form texts from the eight datasets in \tableref{tab:seven_datasets_for_claim_extraction} (Usage = HE), using both \SAFE's method and ours. The datasets were selected to 
have a range of verifiable factual content.
For time and cost efficiency, we only used \SAFE's fact extraction and revision steps (see \appendixref{appendix:issue_in_safe} and Footnote \ref{fn_time_consuming}). To ensure the comparison was independent of the model used, we used GPT-4, paired with \SAFE's and our methods, to extract claims from half of the data points and Claude 3 for the other half.\footnote{All claim extractions are done on April 3\textsuperscript{rd} and 4\textsuperscript{th}, 2024.} For each text, the annotators were asked to choose the claim list that had the most verifiable and the least/no unverifiable content, indicate whether it was hard to choose, and briefly justify their choice. Data preparation and annotation details are in \appendixref{appendix:1stHumanEval}. In total, we collected 360 data points.

\vspace{-5pt}
\paragraph{Results:} The three annotators fully agreed on 99 out of 120 annotated data points, resulting in a Fleiss $\kappa = 0.7662$ (substantial agreement, \citealp{landis1977measurement}).
Of the 360 annotated items, claims extracted by \SAFE~were preferred only 26 times, with 19 of those preferences being marginal. The annotator preferences across the data domains are detailed in \figureref{fig:1stHumanEval}. Notably, our approach is significantly favored even on biography generation.
\vspace{-5pt}
\paragraph{Annotator comments on \SAFE's claims:} The annotators identified three major issues with \SAFE's extraction pipeline.
First, it indiscriminately extracts \textit{everything}, such as subjective content \exampleref{ex.subjective} and personal experience \exampleref{ex.personal}.

\begin{exe}
\fontsize{10}{12}\selectfont
\vspace{-1pt}
    \ex 
    \begin{xlist}
        \ex I am 1000\% better.\label{ex.subjective}

        \ex My grandpa assembled a TV.\label{ex.personal}
    \end{xlist}
\vspace{-1pt}
\end{exe}

\noindent Second, \SAFE~overly decomposes texts, causing meaning overlaps between claims as in \exampleref{ex.overlap}, which can disproportionately affect the final score.


\begin{exe}
\fontsize{10}{12}\selectfont
\vspace{-1pt}
    \ex\label{ex.overlap}
    Longwood House is a place. \\ 
    Longwood House is a Napoleonic Museum. \\ 
    Longwood House is one of the best Napoleonic Museums. \\ 
    Longwood House is one of the best Napoleonic Museums in the world.
\vspace{-1pt}
\end{exe}

\noindent Third, \SAFE~often extracts trivial \exampleref{ex.trivial} or vague claims \exampleref{ex.vague} that do not need to or cannot be verified. 


\begin{exe}
\fontsize{10}{12}\selectfont
\vspace{-1pt}
    \ex
    \begin{xlist}
    \ex 3.2 is a number.\label{ex.trivial} 

    \ex \textit{All My Sons} has key themes.\label{ex.vague}
    \end{xlist}
\vspace{-1pt}
\end{exe}


\subsection{\name's claim extractor only extracts verifiable claims}

To further support that our claim extraction method with GPT-4 extracts only verifiable claims, we applied it to LLM-generated responses to 200 prompts from each domain in \tableref{tab:seven_datasets_for_claim_extraction} (Usage = Dev) and calculated the average ratio of verifiable claims to sentences per response, shown in the VerRatio column of \tableref{tab:seven_datasets_for_claim_extraction}. We observe significant and intuitive differences in this ratio across domains: fact-seeking domains (e.g., FreshBooks) have a higher density of verifiable claims, 
while WritingPrompts' creative story outputs contain almost no verifiable claims with a ratio of 0.03, despite containing the longest responses.\footnote{It has on average 34 sentences per response.} This level of variation shows that our method effectively discriminates verifiable and unverifiable content.


\subsection{Human evaluation on claim verification}\label{sec:claim.verification}

We conducted a human study where three annotators verified claims given search results. The study has three purposes: (1) to understand the feasibility of the task, (2) to see the distribution of the labels in \tableref{tab:claim_veri_def}, and (3) to later judge automatic verifiers by their agreement with human annotations. 


\vspace{-5pt}
\paragraph{Setup:} We sampled 320 GPT-4-extracted claims from model responses to the prompts from the datasets (Usage = Dev) in \tableref{tab:seven_datasets_for_claim_extraction}. Evidence was retrieved as described in \sectionref{subsec:define_evidence_retrieval}. The <claim, evidence list> pairs were split into subsets of 50 for agreement analysis and three subsets of 90, with each annotator doing one. The annotators evaluated each claim on two levels: (1) evidence level: assess if each search result supports, contradicts, or is inconclusive for the claim; (2) claim level: whether the claim is \supported, \contradicted, or \inconclusive~given all the evidence.
\vspace{-5pt}
\paragraph{Human agreement:} The agreement result shows that the verification task is well-defined and feasible. Of the 50 triple-annotated items, 82\% had complete agreement among the annotators, and 14\% had two annotators in consensus. The Fleiss $\kappa$ is $0.7316$ (substantial agreement). An analysis of annotator disagreements is provided in \appendixref{appendix:human_study_on_veri}.

\vspace{-5pt}
\paragraph{Reasons for being inconclusive:} Among the 41 fully agreed items, 15 are \inconclusive. There are two reasons. First, a claim is too general to be verified (e.g., ``A systematic review on sex differences in the reinforcing effects of nicotine was published in Nicotine \& Tobacco Research in 2019.'' without specifying \textit{which} systematic review it was.)
Second, there is no direct mentioning of a part of the claim or no evidence verifies the connection between the parts of a claim (see \tableref{tab:ex_of_inconclu_cases}). Overall, no triple-annotated item is marked as \inconclusive~for the reason that there are both supporting and contradicting search results. 

\vspace{-5pt}
\paragraph{Only over half of the claims are \supported.} We analyzed the distribution of the claim level labels of all annotated items. For the triple-annotated data, we use the majority vote, if there is one, as the final label. Otherwise, the label is \inconclusive. Results in \tableref{tab:claim_veri_def_distribution} show that only 55\% of the claims are \supported. As discussed later in \sectionref{subsec:veriscore_results}, the low supported rate showcases that open-domain claim verification is beyond identifying exact or related terms but requires extensive reasoning to verify the connection between parts of a claim. 

\begin{table}[h!]
\fontsize{6}{7}\selectfont
\centering
\resizebox{0.8\columnwidth}{!}{%
\begin{tabular}{@{}p{2cm}p{0.6cm}p{0.6cm}@{}}
\toprule
\textbf{Label}  & \textbf{Count} & \textbf{\%} \\ \midrule
Claim supported    & 176 & 55\%  \\
Claim contradicted & 9   & 2.8\% \\
Inconclusive (a)   & 128 & 40\%  \\
Inconclusive (b)   & 7   & 2.2\% \\\bottomrule
\end{tabular}%
}
\caption{The distribution of the four labels (\tableref{tab:claim_veri_def}) that can happen in the claim verification step. 
}
\label{tab:claim_veri_def_distribution}
\vspace{-10pt}
\end{table}
\vspace{-5pt}
\paragraph{Top search results are more informative.} We consider a search result informative if it is marked as supporting or contradicting a claim. We analyzed the frequency with which search results were deemed informative. The first five search results show higher utility, with over 30\% being useful---the highest being the first search result at 35.6\%. For search results six to nine, their usefulness percentages range from 27.0\% to 29.2\%. The utility of the last search result drops to only 13.3\%.

\subsection{Automatic verifier}

To find the best performing LLM on the claim verification task, we tested Mixtral-8$\times$22-Instruct-v0.1, Claude 3, GPT-4, and GPT-4o on the human annotated verification data using the binary classification prompt in \tableref{tab:claim_verification_prompt}.\footnote{We also experimented with a ternary classification prompt but the LMs' performance was worse. See \appendixref{appendix:ft_claim_verification}.} We calculated the precision, recall, and $F_1$ on all items as well as separately on the \supported~and \unsupported~items. Results in \tableref{tab:prompted_model_on_320} show that GPT-4o aligns the best with the human performance.
Hence, we use GPT-4o data for fine-tuning an open claim verifier.

\section{Using \name\ to benchmark LM factuality}\label{sec:veriscore_16_models_6_domains}

In this section, we use \name\ to benchmark 16 LMs on 6 long-form fact-seeking domains.\footnote{The results on FreshQA and WritingPrompts are reported in \appendixref{appendix:WP_and_FreshQA} because the former mostly requires short answers and the latter is not fact-seeking.} We first introduce our fine-tuned models for claim extraction and verification, which are used for our large-scale study. Our results highlight the gap between closed and open-weight LMs, which GPT-4o achieving significantly higher \name\ than any open LM. We also note tasks whose \name\ does not correlate well, and conclude with qualitative analysis revealing limitations of \name's verification step.

\subsection{An open-weight \name\ pipeline}\label{subsec:fine-tuned_models}

To facilitate affordable factuality evaluation, we fine-tuned open LMs for a deterministic and cost-efficient \name\ pipeline. We use the few-shot prompting pipeline developed in \sectionref{subsec:claim.extraction} to generate 13403 training data.\footnote{Each data point consists of a claim, search results, and a label.} We experimented with Llama3-8B-Instruct and Mistral-7B-Instruct-v0.2 (henceforth Llama3 and Mistral) as the base models (see \appendixref{appendix:ft_claim_verification} for details of fine-tuning). The benchmark experiments are then performed with the best performing fine-tuned models. The fine-tuned \name\ saves considerable money, making the evaluation process more accessible.\footnote{Evaluating 400 GPT-4o generations in the domains in \tableref{tab:grand_table} using our prompting method cost \$1,038 USD.}



For claim extraction, the fine-tuned Mistral on GPT-4 data achieves the most competitive performance. The model sees the whole prompt and model response and extract claims sentence by sentence. In a quality comparison of 300 pairs of Mistral and GPT-4 extracted claims in \appendixref{appendix:ft_claim_extraction}, the exact match rate is 43.7\% and RougeL is 0.801. For claim verification, a verifier should be equally adept at identifying valid claims as well as recognizing unsupported claims. The fine-tuned Llama3 on GPT-4o data performs the best on the human annotated data in \sectionref{sec:claim.verification}, achieving $F1 = 0.841$ (see \tableref{tab:prompted_model_on_320}). Details of the fine-tuning process and quality analysis are in \appendixref{appendix:ft_claim_extraction} and \ref{appendix:ft_claim_verification}.

\subsection{Data domains and studied LMs}\label{subsec:data_and_lms}

\name\ aims to operate on a wide range of domains. We prompt 16 LMs using prompts from the datasets in \tableref{tab:seven_datasets_for_claim_extraction} and benchmark their factuality. 
The datasets include prompts that require various degree of factual content, from highly fact-dense (e.g., AskHistorians and ELI5) to moderately factual (e.g., ShareGPT). We also collect a dataset FreshBooks that consists of 10 paragraphs from each of 20 non-fictional books in \tableref{tab:new_book_dataset} published between 2023 and 2024. Models are required to generate a continuation of the paragraphs. 

The three largest model families are the GPT, Claude 3, and Mistral/Mixtral models. We also evaluate LMs of various sizes---Qwen1.5-1.8B-Chat, Gemma-2B-it, OLMo-7B-Instruct, Vicuna-7B-v1.5, and DBRX Instruct (132B). Details of the models are in \tableref{tab:simple_12_models_for_response_gen}. To generate model responses for evaluation, the default model hyperparameters were used. The maximum token length was set to 1024. We used 50 prompts per domain.\footnote{The instruction ``Generate a continuation of the following text. The continuation should be objective and factual'' is prepended to the FreshBooks paragraphs.}




\subsection{\name\ results}\label{subsec:veriscore_results}

\begin{table}[t]
\fontsize{6}{8}\selectfont
\resizebox{0.99\columnwidth}{!}{%
\begin{tabular}{@{}p{1.87cm}p{0.2cm}p{0.2cm}p{0.2cm}p{0.2cm}p{0.2cm}p{0.25cm}p{0.4cm}@{}}
\toprule
Dataset & \textbf{LF} & \textbf{Bio} & \textbf{ELI5} & \textbf{AskH} & \textbf{FBs} & \textbf{S.GPT} & \textbf{Avg.} \\
$K$ & (32) & (26) & (21) & (21) & (24) & (11) & \\\midrule
\rowcolor{green!10} Gemma-2B-it & 60.7 & 4.6 & 28.8 & 17.8 & 25.1 & 27.6 & 27.4 \\
\rowcolor{green!10} Mist-7B-Inst-v0.1 & 57.6 & 20.3 & 42.2 & 36.5 & 39.8 & 41.2 & 39.6 \\
\rowcolor{green!10} Vicuna-7B-v1.5 & 63.4 & 23.0 & 51.3 & 39.7 & 39.0 & 43.6 & 43.3 \\
\rowcolor{green!10} Qwen1.5-1.8B-Chat & 70.3 & 14.1 & 57.9 & 45.2 & 52.6 & 49.2 & 48.2 \\
\rowcolor{green!10} OLMo-7B-Inst & 73.4 & 19.4 & 58.8 & 43.2 & 53.7 & 49.4 & 49.6 \\
\rowcolor{green!10} Mist-7B-Inst-v0.2 & 72.0 & 30.0 & 58.8 & 41.2 & 52.4 & \textbf{54.8} & 51.5 \\
\rowcolor{green!10} Mixt-8x7B-Inst-v0.1 & 77.3 & 42.5 & 61.9 & 50.7 & 57.4 & 51.5 & 56.9 \\
\rowcolor{green!10} DBRX-Inst & 75.9 & 46.5 & 61.9 & 49.5 & 60.2 & 48.9 & 57.2 \\
\rowcolor{green!25} Mixt-8x22B-Inst-v0.1 &78.0 & 47.6 & 64.9 & 51.1 & 58.0 & 51.4 & 58.5 \\
\rowcolor{red!10} GPT3.5-turbo-1106 & 64.7 & 38.1 & 42.8 & 40.8 & 32.5 & 42.1 & 43.5 \\
\rowcolor{red!10} Claude-3-Haiku & 79.4 & 37.1 & 58.7 & 43.5 & 49.5 & 44.7 & 52.2 \\
\rowcolor{red!10} Claude-3-Sonnet & 80.7 & 37.6 & 56.2 & 40.7 & 59.3 & 51.7 & 54.4 \\
\rowcolor{red!10} GPT3.5-turbo-0613 & 77.6 & 45.9 & 62.9 & 51.8 & 49.0 & 48.6 & 56.0 \\
\rowcolor{red!10} Claude-3-Opus & 83.6 & 52.7 & 63.4 & 49.8 & 66.4 & 51.6 & 61.2 \\
\rowcolor{red!10} GPT4-0125-preview & 85.9 & 56.4 & 70.7 & 56.6 & 69.7 & 53.5 & 65.5 \\
\rowcolor{red!20} GPT-4o & \textbf{86.7} & \textbf{56.7} & \textbf{71.7} & \textbf{61.4} & \textbf{70.9} & 51.5 & \textbf{66.5} \\\midrule
\textbf{Kendall's $\tau$ w/ Avg.} & 0.78 & 0.83 & 0.82 & 0.73 & 0.73 & 0.56 & 1.00\\\bottomrule
\end{tabular}%
}
\caption{\name\ on 50 responses per LM per dataset (FreshQA and WritingPrompts in \tableref{tab:grand_table}). $K$ is in brackets. Dataset full names are in \tableref{tab:seven_datasets_for_claim_extraction}. Precision and recall are in \tableref{tab:grand_table}. All correlations are statistically significant. GPT-4o is the best\lightred{closed LLM}and Mixt-8x22B-Inst-v0.1 the best\lightgreen{open LLM}.}
\label{fig:model_resp_heatmap}
\vspace{-10pt}
\end{table}

\begin{figure}[t]
    \centering
    \includegraphics[width=0.65\linewidth]{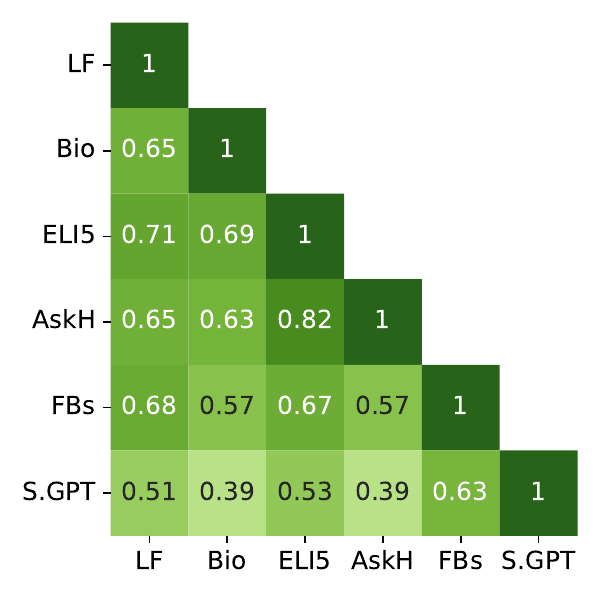}
    \caption{Kendall's $\tau$ correlations of LMs' performance between domains. All correlations are statistically significant. Models' \name\ on dissimilar tasks do not necessarily have high correlation, highlighting the need of using different tasks to assess LMs' factuality.}
    \label{fig:corr_heatmap}
    \vspace{-7pt}
\end{figure}

The factuality performance of the 16 LMs on \name\ is reported in \tableref{fig:model_resp_heatmap}. We tune $K$ in $F_1@K$ for each domain, which is the median number of verifiable claims extracted from each response in each domain from all models.  From the results, we observe the following:
\vspace{-5pt}
\paragraph{Closed models are more factual.} Overall, the GPT models performs better than the Claude 3 models.\footnote{\label{fn:gpt_len}\textnormal{GPT-3.5-turbo-1106 is an exception because the model generates shorter responses than GPT-3.5-turbo-0613, which hurts the recall. On average, GPT-3.5-1106 generates 8.56 sentences per response and GPT-3.5-0613 generates 15.95.}} DBRX-Instruct and the Mixtral models performs competitively to some versions of the GPT and Claude 3 models. The smaller models fall behind on \name, with Gemma-2b-it performing the worse across all domains. The overall trend underscores the correlation of model size and \name\ in long-form fact-seeking outputs.
\vspace{-5pt}
\paragraph{Multiple generation tasks are needed for a comprehensive factuality evaluation.} The Kendall's $\tau$ correlations between LMs' performance in domains in \figureref{fig:corr_heatmap} indicate that LMs' \name\ on two fact-seeking domains (e.g., ELI5 and Biography) do not necessarily correlate well. This suggests that LMs exhibit varying strengths in different domains, highlighting the need for diverse tasks to comprehensively assess LMs' factuality.
\vspace{-5pt}
\paragraph{$F1@K$ favors longer outputs.} 
$F_1@K$ \citep{longfact} considers both factual precision and recall, which improves on measuring factual precision alone \citep{factscore}. A model must generate at least $K$ supported claims per response to achieve perfect recall.
However, for domains that do not require long generations, models that generate short to-the-point outputs will be penalized if other models generate lengthy outputs with auxiliary information (e.g., FreshQA in \appendixref{appendix:WP_and_FreshQA}). It is debatable whether longer responses should always be preferred. They provide more details but, as \citet{factscore} shows, later facts in long responses tend to be less accurate.


\subsection{Qualitative analysis}\label{subsec:qualitative}

This subsection examines \name's performance, highlighting the limitations of decompositional factuality evaluation for generations that are not entity-centric and not formulaic. Two issues are identified: (1) not all claims can be short, and long claims are harder to verify, and (2) search results may be insufficient as expertise or extensive logical reasoning is often needed for verification. 

\subsubsection{Claim complexity increases outside of entity-centric tasks}

Shorter, self-contained claims are desired because they help locate factual errors and are easy to be verified as employed by \factscore~for biography.
However, claims extracted from other fact-seeking generations are often long.\footnote{To confirm this, we randomly sampled 200 claims from \citet{factscore}'s model extracted claims and 200 claims extracted by GPT-4 from GPT-4 generated ELI5 responses. On average, \factscore's claims have 7 words, with the longest one having 18 words. In contrast, the ELI5 claims on average have 12 words, with the longest one having 25 words.} While some long claims could be split at conjunctions like \textit{and} or \textit{or}, this does not significantly shorten claims with inherently long core content, as seen in \exampleref{ex.crusaders}.

\begin{exe}
\fontsize{10}{12}\selectfont
\vspace{-2pt}
    \ex\label{ex.crusaders}
    Travelers \textit{and} crusaders during the medieval period depended on established infrastructure to secure clean \textit{and} consistent sources of water.
\vspace{-18pt}
\end{exe}

\noindent Occasionally, shorter claims can be extracted from a longer one, as the bracketed content in \exampleref{ex.Chuck}. However, verifying the shorter claims does not mean the longer one is verified because of \textit{solidified}.

\begin{exe}
\fontsize{10}{12}\selectfont
\vspace{-1pt}
    \ex\label{ex.Chuck}
    [Chuck Norris's victory in the 1968 World Full-Contact Karate Championships] \textit{solidified} [his reputation as one of the best martial artists in the world].
\vspace{-2pt}
\end{exe}

\noindent For these reasons, long and complex claims are likely to be marked as \inconclusive. 

\subsubsection{Google Search may be insufficient for complex claims}\label{subsubsec:search_result_insufficient}

To understand what types of claims are supported and unsupported by Google Search snippets, we examine 80 claims from GPT-4o generated ELI5 and FreshBooks responses, along with their search and verification results. Half of these claims were verified as supported and the other half were not.\footnote{More \unsupported~claim examples are in \appendixref{appendix:veriscore_inconclusive}.}
\vspace{-18pt}
\paragraph{The supported claims resemble encyclopedic writing.} The content is supported by search results via semantic or string match, as in \exampleref{ex.indigenous}.

\begin{exe}
\fontsize{10}{12}\selectfont
\vspace{-1pt}
    \ex\label{ex.indigenous}
    Indigenous women in Australia were not fully enfranchised until much later.\footnote{Search snippet: In Australia, Indigenous women were not enfranchised until 1962, six decades after non-Indigenous women were able to vote. \href{https://www.pewresearch.org/short-reads/2020/10/05/key-facts-about-womens-suffrage-around-the-world-a-century-after-u-s-ratified-19th-amendment/}{(link)}}
\vspace{-5pt}
\end{exe}
\vspace{-5pt}
\paragraph{The unsupported claims do not have direct contradicting evidence.} They are unsupported because there is no direct mention of (parts of) the claims or the connection between the parts of the claims. Example \exampleref{ex.tiger} is judged as \unsupported~because there is no mention of the Meiji era and stuffed tigers occurring together in the search results.\footnote{After extensive search, we did not find any supporting or contradicting evidence to the claim. 
} Snippets do not offer enough background for such reasoning. Expertise or more sophisticated search is needed to verify/falsify such claims.

\begin{exe}
\fontsize{10}{12}\selectfont
\vspace{-2pt}
    \ex\label{ex.tiger}
    Japanese people encountered tigers in the form of stuffed animals before the Meiji era. 
\vspace{-6pt}
\end{exe}

\vspace{-5pt}
\paragraph{Some unsupported claims require extensive supporting evidence.} This happens the most often in FreshBooks when a claim encapsulates aspects like someone's achievements or historical movements, as in (\ref{ex.marshall}-\ref{ex.germany}). Such content might not be directly mentioned in search results but need to be inferred from a large body of documents.

\vspace{-1pt}
\begin{exe}
\fontsize{10}{12}\selectfont
    \ex\label{ex.marshall}
    Marshall’s leadership and strategic acumen ensured the maneuver was carried out flawlessly during a field maneuver in the Philippines. 

    \ex\label{ex.germany}
    Germany is maintaining its competitive edge in a rapidly changing global landscape. 
\end{exe}
\vspace{-1pt}

\noindent In sum, with the current system, it is hard to decide whether an \unsupported~claim is hallucinated because it is beyond what reasoning over search snippets can achieve. This indicates the need to move beyond semantic or string matching for verification as they fail to uncover possible hallucination.

\section{Related work}
Our work builds on prior research in claim verification and long-form factuality evaluation. 
Users rely on the accuracy of LLM-generated content, yet LLMs often produce unreliable information \citep{maynez-etal-2020-faithfulness, xu-etal-2023-critical, huang2023survey, rawte2023survey}. Research has thus focused on enhancing factual precision \citep{lin2024flame} and identifying inaccuracies.

\paragraph{Factual error detection:} Prior research targets error detection in individual sentences \citep{mihaylova-etal-2019-semeval, wadden-etal-2020-fact, shaar-etal-2022-role}. FEVER \citep{thorne-etal-2018-fever} features synthesized incorrect sentences from Wikipedia. FEVEROUS \citep{aly-etal-2021-fact} and \textsc{AVeriTeC} \citep{schlichtkrull2023averitec} build on FEVER but remained limited to sentence-level facts. At the paragraph level, \citet{li-etal-2023-halueval} test LLMs' detection of synthesized factual errors but do not locate the errors. 

\paragraph{Long-form factuality evaluation:} Detecting factual errors in a long-form text at once is hard \citep{li-etal-2023-halueval}. Decomposing a piece of long-form text into shorter sentences or search queries for factuality evaluation is commonly implemented in previous works \citep{kamoi-etal-2023-wice, rarr, factcheck, factscore, chern2023factool, wanner2024closer, longfact, guan-etal-2024-language, chen-etal-2024-complex}. The decomposition helps locate factual errors and offers a fine-grained estimate of models' factuality \citep{factscore}. Factuality evaluation often requires world knowledge, which can be achieved by employing retrieval \citep{ram-etal-2023-context, vu2023freshllms}. It is commonly used in factual error detection \citep{factscore, longfact, thorne-etal-2018-fever} and helps evaluation by providing up-to-date knowledge. The overall evaluation pipeline helps generate post hoc citations and iteratively improve model generations' factuality, which improves models' trustworthiness model \citep{huang-chang-2024-citation, ye-etal-2024-effective}.

\section{Conclusion and future work}\label{sec:conclusion}

We propose \name, a factuality metric that focuses exclusively on \emph{verifiable} claims. Human evaluations validate that \name\ is more effective than existing metrics for diverse long-form generation tasks that contain both verifiable and unverifiable content. We open-source both a closed- and open-weight implementation of \name, with the latter's performance approaching that of the former.
Finally, we notice that complex claims (e.g., not entity-centric or formulaic) are challenging to verify against search results. We hope that future work will improve on this aspect to develop more robust factuality metrics.

\section*{Limitations}

We acknowledge further limitations of the current work and the decompositional approach below.

First, formally defining verifiable claims poses a significant challenge. Although our definition advances beyond the concept of \textit{atomic facts} \citep{factscore}, it remains a working definition rather than a formal one. For instance, consider the sentence \exampleref{ex.Chuck}: it is problematic to  determine whether it describes a single state of "solidifying" or encompasses one event, "Chuck Norris's victory in ... Championships," along with two states, "solidifying" and "as one of the best ... in the world,". 
We hope future studies can improve on this.

Second, the decomposition method is slow. With one RTX8000 GPU, it takes about 4 hours to extract claims from 400 GPT-4o responses without parallelization. The reason is that \name\ uses a sliding window to scan through a model response. In our experiments, each response on average has 40 sentences (20 sentences on average if excluding WritingPrompts responses). For verification, it takes about two hours to verify 10k claims. Future work can aim for a claim extractor that works without a sliding window to speed up the claim extraction.



Third, for model responses that are extremely infactual (e.g., WritingPrompts in \appendixref{appendix:WP_and_FreshQA}), our claim extractor might still extract a small amount of unverifiable claims. However, we contend that the percentage of factual content in creative writing in response to fictional premises is less concerned than in the fact-seeking domains. Hence, we do not consider this as a major concern of \name.

Fourth, in the current work, we did not search exhaustively for the best hyperparameters for fine-tuning the open-source claim extractor and verifier. It is possible that, after searching, a better performance can be achieved. However, it is resource-intensive and time-consuming.

\section*{Ethics Statement}

Our project aimed to minimize the computational cost by using LoRA \citep{hu2022lora} for efficient model fine-tuning. For the annotation work, an IRB review was exempted. By signing a data conset, each annotators agreed on the annotated data being used for scientific research and published. No personally identifiable information was collected. We paid annotators \$18 per hour. Additional bonus were paid for reasonable extra time spent.

\section*{Acknowledgement}

We extend our special gratitude to Kalpesh Krishna, who extensively discussed the project details with us and offered invaluable insights. We extend gratitude to the Upwork annotators for their hard work, and to members from the UMass NLP lab for their feedback. This project was partially supported by awards IIS-2202506, IIS-2046248, and IIS-2312949 from the National Science Foundation (NSF).

\bibliography{anthology,custom}

\appendix

\section{Weaknesses in \factscore~and SAFE}\label{appendix:issue_in_safe}

In \sectionref{subsubsec:issues_in_safe}, we pointed out four major issues in SAFE's claim extraction pipeline. Details of the issues are provided in this appendix section.


First, for claim extraction, aside from prepending a brief task description to \factscore's prompt, SAFE does not make other modifications. Consequently, the prompt only focuses on biography. 

Second, SAFE's extraction pipleine is multi-step. Because \factscore~extracts claims by sentence without context, it cannot resolve references. This limitation is not an issue for \factscore~because each claim is verified against one predefined Wikipedia article. However, SAFE uses Google Search and thus must resolve all vague references. SAFE addresses this by deploying claim revision which prompts a language model once for each claim to revise vague references. Following that, a language model reviews each claim again to decide whether they are worth checking. The entire pipeline adds significant processing time and cost.

Third, the relevance check step negatively impacts evaluation. \citet{longfact} justifies this step with an example in their Figure 1---when asked about the Eiffel Tower, a model generates \textit{The Nile River is in Egypt}. First of all, such behaviour is not observed in our experiments. Second, we applied SAFE's extraction pipeline to five texts and examined which claims were removed. It turns out that 11\% of 211 claims were removed, of which 58\% were actually relevant. The remaining 42\% were either tautologies or not claims and should not have been extracted.\footnote{For example, \textit{Castello Maniace is Castello Maniace.} is a tautology; \textit{As always, there is some disclaimer.} is not a verifiable claim.} \tableref{tab:safe_relevance} provides an example of SAFE removing a relevant claim.

\begin{table*}[t]
\fontsize{9}{11}\selectfont
\centering 
\resizebox{\textwidth}{!}{%
\begin{tabular}{@{}p{16cm}@{}}
\toprule
\multicolumn{1}{c}{\textbf{\scalebox{1.1}{SAFE's relevance check identifies relevant claim as irrelevant.}}}
\\\midrule

\textbf{Question}: At their peak, what did the insides of the most beautifully decorated castles look like? Today, castles seem to just be giant fortresses but I would like to know how they looked when they were fully furnished. How were they decorated? What treasures were stored there? Are there a few castles that were especially beautiful?

\textbf{Human response}: It is quite a broad subject because castles varied quite a lot depending on location, time of construction and wealth of the constructor; u/valkine talked about Caenarfon Castle \href{https://www.reddit.com/r/AskHistorians/comments/26vvmz/were_castles_in_medieval_england_typically/}{(link)} specifically in another question \href{https://www.fortementein.com/wp-content/uploads/2021/03/p-le9102-1280x720.jpg}{(link)} is a part of the inside of \colorbox{yellow}{Castello Maniance in Siracusa, Italy. It was built from 1232 to 1239} during a large castle-construction effort by Emperor Frederick II.  I do find it particularly beautiful but this doesn't really say much about what other castles looked like.

\textbf{Extracted claim}: Castello Maniance in Siracusa, Italy was built from 1232 to 1239.\\

\textbf{Authors' note}: Although the human answer does not answer all parts of the question, the content that is deemed as irrelevant by SAFE is actually pointing to a castle that is relevant to answering the question.\\\midrule

\end{tabular}%
}
\caption{An example illustrating SAFE's relevance assessment does not work as expected.}
\label{tab:safe_relevance}
\end{table*}

Fourth, there is no guarantee that SAFE works across domains. Despite being applied to 38 fact-seeking topics, SAFE's performance is only evaluated on \factscore's biography data. Among the 38 topics, SAFE is solely applied to model outputs that responses to object-related prompts. Six topics mostly contain biography questions (i.e., \textit{Who is}).\footnote{The six topics are: celebrities, jurisprudence, mathematics, medicine, philosophy, and sociology.} Some topics (e.g., sports) contain only \textit{who}, \textit{what}, and \textit{can you tell me about} questions, making them fact-dense and entity-centric. A human study in \sectionref{subsec:1stHumanEval} confirms that SAFE falls short in less entity-centric domains.



\section{Formal definition of claim verification}\label{appendix:formal_def_of_verification}

In \sectionref{subsec:define_claim_verification}, we described the definition of the four possible scenarios that can happen when verifying a claim with respect to evidence. We give a formal definition of such scenarios in \tableref{tab:claim_veri_def}.

\begin{table*}[!ht]
\fontsize{8}{10}\selectfont
\centering
\resizebox{0.8\textwidth}{!}{%
\begin{tabular}{@{}p{2.2cm}p{7.8cm}@{}}
\toprule
\textbf{Scenario}  & \textbf{Description}  \\ \midrule
Claim supported      & $\forall p \in c.[\exists e \in E_c. \texttt{support}(e,p) \wedge \neg\exists e \in E_c. \texttt{contradict}(e,p)]$  \\[1.5mm]
Claim contradicted   & $\exists p \in c.[\neg\exists e \in E_c. \texttt{support}(e,p) \wedge \exists e \in E_c. \texttt{contradict}(e,p)]$ \\[1.5mm]
Inconclusive (a) & $\exists p \in c.[\neg\exists e \in E_c. \texttt{support}(e,p) \wedge \neg\exists e \in E_c. \texttt{contradict}(e,p)]$ \\[1.5mm]
Inconclusive (b) & $\exists p \in c.[\exists e \in E_c. \texttt{support}(e,p) \wedge \exists e \in E_c. \texttt{contradict}(e,p)]$\\\bottomrule
\end{tabular}%
}
\caption{Four scenarios that can happen in the claim verification step. \texttt{support}$(a,b)$ and \texttt{contradict}$(a,b)$ are two transitive predicates such that $\neg$\texttt{support}$(a,b) \neq$ \texttt{contradict}$(a,b)$ and  $\neg$\texttt{contradict}$(a,b) \neq$ \texttt{support}$(a,b)$.
}
\label{tab:claim_veri_def}
\end{table*}

\section{Claim extraction prompts}
\label{appendix:our_claim_extraction_prompts}


We developed two claim extraction prompts: one for question-answering (QA) type of input data, and the other for non-QA data. For evaluating model outputs, the QA prompt is generally applicable with the prompt being the question. The non-QA prompt is used for cases where neither a question nor a prompt is available.

What is common in the two prompts is a sliding window for claim extraction. Each window has the format \texttt{(context1 = 0-3 sentence) <SOS>Sentence to be focused on<EOS> (context2 = 0-1 sentence)}. The goal is to extract claims from the sentences marked by \texttt{SOS} and \texttt{EOS} while using the information in \texttt{context1} and \texttt{context2} to make the claims self-contained.

What is different in the two prompts is that for non-QA-type of inputs, we always prepend the first sentence of a paragraph to \texttt{context1} if the paragraph is longer than five sentences; for QA-type of inputs, we always prepend the question to \texttt{context1}. This is based on the observation that, when answering a question or when an answer gets long, people might take the information in the question or previous sentences for granted and do not refer to an entity using its full name. Adding the question or the first sentence of a paragraph into \texttt{context1} can help a model better recover time, location, and person references in a claim. 

The prompts are given in \tableref{tab:non-QA-claim-extraction} and \tableref{tab:QA-claim-extraction}

\begin{table*}[ht]
\fontsize{10}{12}\selectfont
\centering 
\resizebox{\textwidth}{!}{%
\begin{tabular}{@{}p{20cm}@{}}
\toprule
\multicolumn{1}{c}{\textbf{\scalebox{1.1}{Prompt for Extracting Verifiable Claims from Non-Question-Answering Type of Inputs}}}
\\\midrule

You are trying to verify how factual a piece of text is. To do so, you need to break down a sentence and extract as many fine-grained facts mentioned in the sentence as possible. Each of these fine-grained facts should be verifiable against reliable external world knowledge (e.g., via Wikipedia). Any story, personal experiences, hypotheticals (e.g., ``would be" or subjunctive), subjective statements (e.g., opinions), suggestions, advice, instructions, and other such content should not be included in the list. Biographical, historical, scientific, and other such texts are not personal experiences or stories. You should extract verifiable facts from them. Each fact should also be describing either one single event (e.g., ``Nvidia is founded in 1993 in Sunnyvale, California, U.S.") or single state (e.g., ``
\colorbox{black}{ABCDEFG} has existed for 161 years.") with necessary time and location information. Quotations should be extracted verbatim with the source when available. Listed references should be ignored. \newline \newline Extract fine-grained facts from the sentence marked between <SOS> and <EOS>. You should focus on the named entities and numbers in the sentence and extract relevant information from the sentence. Other sentences are only context for you to recover pronouns, definite phrases (e.g., ``the victims" or ``the pope"), and so on. Each fact should be understandable on its own and require no additional context. This means that all entities must be referred to by name but not pronoun. Use the name of entities rather than definite noun phrases (e.g., 'the teacher') whenever possible. If a definite noun phrase is used, be sure to add modifiers (e.g., a embedded clause, a prepositional phrase, etc.). Each fact must be situated within relevant temporal and location whenever needed. Keep each fact to one sentence with zero or at most one embedded clause. You do not need to justify what you extract. \newline \newline If there is no verifiable fact in the sentence, please write ``No verifiable claim."\\[5mm]

Here are some examples: \newline \newline Text: The sweet potato or sweetpotato (Ipomoea batatas) is a dicotyledonous plant that belongs to the bindweed or morning glory family, Convolvulaceae. <SOS>Its large, starchy, sweet-tasting tuberous roots are used as a root vegetable.<EOS> The young shoots and leaves are sometimes eaten as greens. \newline Sentence to be focused on: Its large, starchy, sweet-tasting tuberous roots are used as a root vegetable. \newline Facts: \newline - Sweet potatoes' roots are large. \newline - Sweet potatoes' roots are starchy.\newline - Sweet potatoes' roots are sweet-tasting.\newline - Sweet potatoes' roots are tuberous.\newline - Sweet potatoes' roots are used as a root vegetable. \newline \newline Text: Garnett had spent well over a decade with the Minnesota Timberwolves, and while he stayed loyal to that team, he found little success there. <SOS>When he said “you can’t get your youth back,” he meant it - because from a human standpoint, had he been able to apply his talents somewhere else, NBA history might have been different.<EOS> \newline Sentence to be focused on: When he said “you can’t get your youth back,” he meant it - because from a human standpoint, had he been able to apply his talents somewhere else, NBA history might have been different. \newline Facts: \newline - Kevin Garnett said ``you can’t get your youth back." \newline \newline Text: I (27f) and my fiance ``Leo" (27m) decided to let my FSIL ``Maya" (32f) stay at our house because she needed space from her husband due to some relationship struggles they're having. Leo and I had gotten wedding cake samples from an expensive bakery specializing in wedding cakes. We planned to test them along with Maya after we finished up some other wedding plans yesterday. <SOS>However, when I came home from work to see Leo yelling at Maya, the box the samples came in wide open on the living room table, and Maya arguing with him.<EOS> I asked what was happening, and Leo angrily told me that while we were both at work, Maya had some friends over and they ended up eating almost all of our cake samples. \newline Sentence to be focused on: However, when I came home from work to see Leo yelling at Maya, the box the samples came in wide open on the living room table, and Maya arguing with him.\newline Facts:\newline No verifiable claim.\newline\newline \ldots~<Total of 13 Examples>~\ldots\\[5mm]

Extract *verifiable atomic* facts. \newline \newline Text: \{\texttt{sliding window}\}\newline Sentence to be focused on: \{\texttt{sentence}\}\newline Facts:\\
\bottomrule

\end{tabular}%
}
\caption{Claim extraction prompt for non-question-answering type of inputs. The sliding window follows the template \texttt{(context1 = 0-3 sentence) <SOS>Sentence to be focused on<EOS> (context2 = 0-1 sentence)}. If the paragraph from which the sentence is taken is longer than five sentences, the first sentence of the paragraph is always prepended before \texttt{context1}. Marked out content will be uncovered after the review process.}
\label{tab:non-QA-claim-extraction}
\end{table*}

\begin{table*}[ht]
\fontsize{10}{12}\selectfont
\centering 
\resizebox{\textwidth}{!}{%
\begin{tabular}{@{}p{20cm}@{}}
\toprule
\multicolumn{1}{c}{\textbf{\scalebox{1.1}{Prompt for Extracting Verifiable Claims from Question-Answering Type of Inputs}}}
\\\midrule

You are trying to verify how factual a response to a question or request is. To do so, you need to break down a sentence and extract as many fine-grained facts mentioned in the response. Each of these fine-grained facts should be verifiable against reliable external world knowledge (e.g., via Wikipedia). Any story, personal experiences, hypotheticals (e.g.,``would be" or subjunctive), subjective statements (e.g., opinions), suggestions, advice, instructions, and other such content should not be included in the list. Biographical, historical, scientific, and other such texts are not personal experiences or  stories. You should extract verifiable facts from them. Each fact should also be describing either one single event (e.g., ``Nvidia is founded in 1993 in Sunnyvale, California, U.S.") or single state (e.g., ``
\colorbox{black}{ABCDEFG} has existed for 161 years.") with necessary time and location information. Quotations should be extracted verbatim with the source when available. Listed references should be ignored. \newline Extract fine-grained facts from the sentence between <SOS> and <EOS>. You should focus on the named entities and numbers in the sentence and extract relevant information from the sentence. Do not extract claims from the question. The question and other sentences are only context for you to recover pronouns, definite phrases (e.g., ``the victims" or ``the pope"), and so on. Each fact should be understandable on its own and require no additional context. This means that you need to always related the extracted claims to the question. This also means that all entities must be referred to by name but not pronoun. Use the name of entities rather than definite noun phrases (e.g., `the teacher') whenever possible. If a definite noun phrase is used, be sure to add modifiers (e.g., a embedded clause, a prepositional phrase, etc.). Each fact must be situated within relevant temporal and location whenever needed. Keep each fact to one sentence with zero or at most one embedded clause. You do not need to justify what you extract.  \newline If there is no verifiable fact in the sentence, please write ``No verifiable claim."\\[5mm]

Here are some examples: \newline \newline Question: What NASA programs would support our college in starting a robotics program? \newline Response: NASA has several programs that can support colleges in starting a robotics program. Here are a few: \newline <SOS>1. NASA Robotics Alliance Project (RAP): This program provides educational resources and support for robotics teams, including college-level teams, that are participating in NASA robotics competitions.<EOS> \newline 2. NASA Minority University Research and Education Project (MUREP): This program provides funding and resources for colleges and universities with a significant minority student population to develop research and education programs in STEM fields, including robotics. \newline 3. NASA's Robotics Education Project: This project provides robotics education materials and resources for educators, including college-level educators, to use in their classrooms. \newline 4. NASA's Space Technology Mission Directorate (STMD): This directorate funds research and development in advanced technologies, including robotics, that can support NASA's mission to explore space. \newline Sentence to be focused on: 1. NASA Robotics Alliance Project (RAP): This program provides educational resources and support for robotics teams, including college-level teams, that are participating in NASA robotics competitions. \newline Facts: \newline - NASA has a program called NASA Robotics Alliance Project (RAP). \newline - NASA Robotics Alliance Project provides educational resources for robotics teams. \newline - NASA Robotics Alliance Project provides supports for robotics teams. \newline - NASA Robotics Alliance Project provides supports for college-level teams that are participating in NASA robotics competitions.\newline\newline Question: How do trees know when to stop growing?  \newline Thanks everyone i learned a lot more about trees.(: \newline Response: <SOS>Ah yes, tomatoes, this is a big problem with tomato plants.<EOS> \newline Sentence to be focused on: Ah yes, tomatoes, this is a big problem with tomato plants. \newline Facts: \newline No verifiable claim.\newline \newline  \ldots~<Total of 10 Examples>~\ldots\\[5mm]

Extract *verifiable atomic* facts. \newline \newline \{\texttt{sliding window}\}\newline Sentence to be focused on: \{\texttt{sentence}\}\newline Facts:\\
\bottomrule

\end{tabular}%
}
\caption{Claim extraction prompt for question-answering type of inputs. The sliding window consists of the question and \texttt{(context1 = 0-3 sentence) <SOS>Sentence to be focused on<EOS> (context2 = 0-1 sentence)}. Marked out content will be uncovered after the review process.}
\label{tab:QA-claim-extraction}
\end{table*}

\section{Human evaluation of claim extractions by our prompt and SAFE}
\label{appendix:1stHumanEval}

To verify the effectiveness of our proposed claim extraction method, we conducted a human evaluation of pair-wise comparison between claims extracted by our prompts and SAFE's. We hired three experienced data annotators on Upwork\footnote{\url{https://www.upwork.com/}}.

To prepare the data for evaluation, we sampled 15 data points from each dataset in \tableref{tab:seven_datasets_for_claim_extraction}. We used the first four datasets as non-QA datasets and the others as QA datasets. Each data point was truncated to 300 white-space-separated words at the sentence boundary. Each annotator was asked to annotate the same set of 120 sampled data. 

\begin{figure*}[!ht]
    \centering
    \includegraphics[scale=0.63]{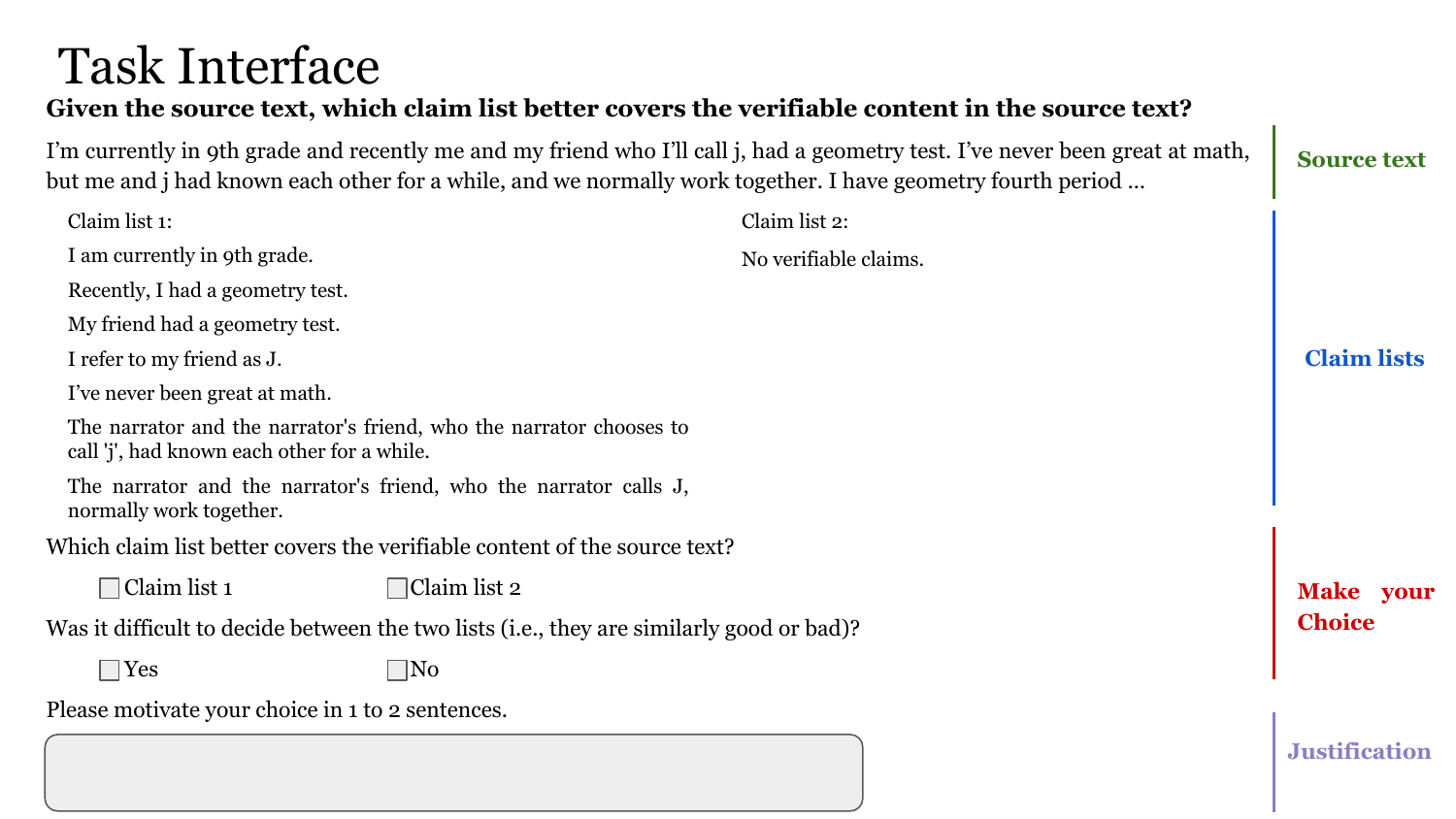}
    \caption{The interface design of the human evaluation described in \sectionref{subsec:1stHumanEval} and \appendixref{appendix:1stHumanEval}. The interface consists of four parts. Source text: the text from which claims are extracted. Claim lists: Two claim lists extracted by our prompt and SAFE respectively. The order of the two lists are randomized. Decisions: annotators indicate here which claim list is better and whether it is hard to choose between the two. Justification: annotators should briefly explain why they choose one list over the other.}
    \label{fig:1stHumanEval_interface}
\end{figure*}

The evaluation was conducted on the open-source data labeling platform Label Studio~\citep{LabelStudio}. The task interface is given in \figureref{fig:1stHumanEval_interface}. Before the task begins, each annotator needs to read through the instructions of the task.\footnote{The instructions are on \href{https://docs.google.com/presentation/d/1qDcOyzQAgj0TY-rRTdHXAjVuVqKpf7BuWx8cFXCz8I0/edit?usp=sharing}{Google slides}.} We estimated the annotation task to take approximately two hours to complete. Therefore, each annotator was compensated at a rate of \$$15$ per hour.

\begin{figure*}[!ht]
    \centering
    \includegraphics[scale=0.42]{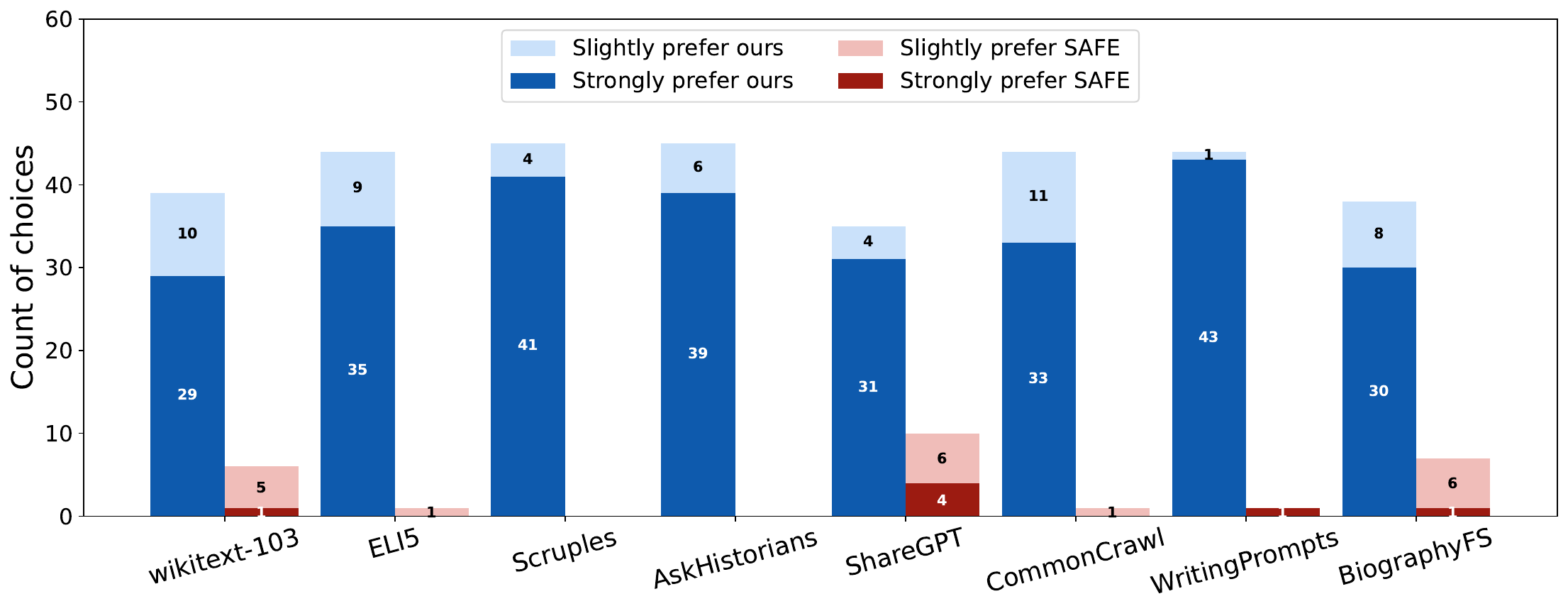}
    \caption{Results of the pair-wise performance comparison between our one-step extraction prompts and SAFE's two-step extraction pipeline. The claims extracted by our prompts are overwhelmingly preferred by the annotators across all eight domains. The dark area in each bar indicates that the annotators strongly preferred one choice over the other. The light area represents slight preference. The numbers are aggregated over three annotators.}
    \label{fig:1stHumanEval}
\end{figure*}

\figureref{fig:1stHumanEval} depicts the human preference in each domain of data in \tableref{tab:seven_datasets_for_claim_extraction}. Among the 360 annotated data points, the claims extracted by SAFE are only preferred 26 times by the three annotators in total, among which 19 were chosen hesitatingly, as indicated by the light red color in \figureref{fig:1stHumanEval}.

\section{Human study on claim verification}\label{appendix:human_study_on_veri}

This appendix section provides supplementary details to \sectionref{sec:claim.verification}.

\subsection{Detailed examples of human verification}

In this section, we provide detailed examples from our human study on the claim verification task in \sectionref{sec:claim.verification}. \tableref{tab:ex_of_inconclu_cases} presents the examples of the annotation items whose claim was labeled as \inconclusive~by all annotators. 

\begin{table*}[ht!]
\fontsize{8}{10}\selectfont
\resizebox{\textwidth}{!}{%
\begin{tabular}{@{}p{\textwidth}@{}}
\toprule
\multicolumn{1}{c}{\textbf{The claim is too general to be verified.}}\\\midrule

\textbf{Example Claim 1}:\newline A wooden spoon creates a small gap between the pot and the spoon.\\

\textbf{Example Claim 2}:\newline Martha had passed away.\\

\textbf{Example Claim 3}:\newline A systematic review on sex differences in the reinforcing effects of nicotine was published in the journal Nicotine \& Tobacco Research in 2019.\\\midrule

\multicolumn{1}{c}{\textbf{A part of the claim is not mentioned in the evidence,}}\\

\multicolumn{1}{c}{\textbf{or no evidence confirms the relationship between the parts in a claim---Inconclusive (a) Case}}\\\midrule

\textbf{Reasoning}:\newline 
(1) Only the search result 7 hints that there might be a Persuasive Technology Lab at Stanford University.\newline (2) No evidence mentions this lab aims to create positive behavior change.\\

\textbf{Claim}\newline The intention of Stanford University's \colorbox{yellow}{Persuasive Technology Lab} was \colorbox{yellow}{to create positive behavior change}.\\

\textbf{Evidence}\newline Search result 1\newline Title: Behavior Design Lab - Stanford University\newline Content: Behavior Design is a new approach to understanding human behavior and how to design for behavior change. Based on the work of Dr. BJ Fogg , Behavior Design ...\newline Link: https://behaviordesign.stanford.edu/\newline Search result 2\newline Title: About Us - Behavior Design Lab - Stanford University\newline Content: Our lab's overall mission is this: Teach good people how human behavior works so they can create solutions that effectively increase health, boost happiness, ...\newline Link: https://behaviordesign.stanford.edu/about-us\newline Search result 3\newline Title: Building Habits: The Key to Lasting Behavior Change\newline Content: “Habits are easier to form than most people think,” he says, “If you do it in the right way.” As the founder and director of Stanford's Behavior ...\newline Link: https://www.gsb.stanford.edu/insights/building-habits-key-lasting-behavior-change\newline Search result 4\newline Title: The Ethical Use of Persuasive Technology - Behavior Design Lab\newline Content: While our research has moved on from persuasive technology to focus on designing for healthy behavior change, we believe it is important to continue to ...\newline Link: https://behaviordesign.stanford.edu/ethical-use-persuasive-technology\newline Search result 5\newline Title: Fiddling With Human Behavior - WIRED\newline Content: Researchers at Stanford are studying technology designed to persuade people to change the way they think or act.\newline Link: https://www.wired.com/2000/03/fiddling-with-human-behavior/\newline Search result 6\newline Title: BJ Fogg - Behavior Design Lab - Stanford University\newline Content: BJ wrote a seminal book, Persuasive Technology: Using Computers to Change What We Think and Do , about how computers can be designed to influence attitudes ...\newline Link: https://behaviordesign.stanford.edu/people/bj-fogg\newline Search result 7\newline Title: How Stanford Profits Off Addiction\newline Content: Back in 1998, one of Stanford's eccentric social scientists, B.J. Fogg, founded the Persuasive Technology Lab to research how tech products ...\newline Link: https://stanfordreview.org/how-stanford-profits-tech-addiction-social-media/\newline Search result 8\newline Title: Tech companies use “persuasive design” to get us hooked ... - Vox\newline Content: Big tech now employs mental health experts to use persuasive technology, a new field of research that looks at how computers can change the way ...\newline Link: https://www.vox.com/2018/8/8/17664580/persuasive-technology-psychology\newline Search result 9\newline Title: Stanford Behavior Design Lab - Wikipedia\newline Content: The Stanford Behavior Design Lab is a research organization advancing behavior change methods and models based at Stanford University. Founded in 1998 and ...\newline Link: https://en.wikipedia.org/wiki/Stanford\_Behavior\_Design\_Lab\newline Search result 10\newline Title: How to create new good habits, according to Stanford ... - Quartz\newline Content: To create a real lifelong habit, the focus should be on training your brain to succeed at a small adjustments, then gaining confidence from that ...\newline Link: https://qz.com/877795/how-to-create-new-good-habits-according-to-stanford-psychologist-b-j-fogg \\\bottomrule

\end{tabular}%
}
\caption{Examples of the annotaton items whose claim was labeled as \inconclusive~by all three annotators. }
\label{tab:ex_of_inconclu_cases}
\end{table*}

\begin{table*}[!ht]
\fontsize{8}{10}\selectfont
\resizebox{\textwidth}{!}{%
\begin{tabular}{@{}p{\textwidth}@{}}
\toprule
\multicolumn{1}{c}{\textbf{Annotator mistake}}\\\midrule

\textbf{Explanation}: Two annotators chose the \inconclusive~label for this claim but one chose \supported~based on one search result as given below. This is an annotator mistake because the name Luis Guillermo Rivera is not mentioned in the evidence.\\

\textbf{Claim}: \colorbox{yellow}{Luis Guillermo Rivera} has written literary criticism. \\

\textbf{Evidence}:\newline 
Search result 5\newline 
Title: I Write with Words That Have Shadow but Don't Shelter
\newline Content: Born in Tumeremo, Bolívar, in 1933, Venezuelan writer Guillermo Sucre is also an essayist, translator, literary critic, and educator. A ...\newline Link: https://www.worldliteraturetoday.org/blog/poetry/i-write-words-have-shadow-dont-shelter-guillermo-sucre\\\midrule

\multicolumn{1}{c}{\textbf{Ambiguity in the interpretation of the claim and evidence}}\\\midrule

\textbf{Explanation}: Two annotators chosen the \supported~label but one chose \inconclusive. The annotator's comment states:``I chose inconclusive as the claim is `can be chosen' and all the results are that they potentially `could' be chosen, not that they actually CAN.'' \\

\textbf{Claim}: Traits that \colorbox{yellow}{can} be chosen include eye color, hair color, intelligence, and athletic ability.\\
\textbf{Evidence}:\newline 
Search result 2\newline
Title: [PDF] Sex Selection, Genetic Analysis, and Designer Babies\newline
Content: In theory, parents \colorbox{yellow}{could} also select embryos on the basis of eye color, hair color, or any other genetic trait.\newline
Link: https://med.nyu.edu/departments-institutes/population-health/divisions-sections-centers/medical-ethics/sites/default/files/medical-ethics-sex-selection-genetic-analysis.pdf \\\midrule

\multicolumn{1}{c}{\textbf{The claim contains an unclear referent.}}\\\midrule

\textbf{Explanation}: The annotators did not agree on this item at all. It is probably because there are multiple people with the name Jessica Barboza, making it hard to make a decision.\\

\textbf{Claim}: Jessica Barboza was born in São Paulo, Brazil.\\

\textbf{Evidence}:\newline
Search result 1\newline
Title: Jessica Barboza - Wikipedia\newline
Content: Jessica Barboza. Born. Jessica Cristina Barboza Schmidt. (1987-08-14) 14 August 1987 (age 36). Maracaibo, Zulia, Venezuela. Height, 1.79 m (5 ft 10+1⁄2 in).\newline
Link: https://en.wikipedia.org/wiki/Jessica\_Barboza\newline
Search result 3\newline
Title: Jessica Barboza - Age, Family, Bio | Famous Birthdays\newline
Content: Style blogger and makeup guru known for her Peace and Vogue blog and YouTube channel. The blossoming beauty maven has gained a following of more than 550,000 ...\newline
Link: https://www.famousbirthdays.com/people/jessica-barboza.html\newline
Search result 5\newline
Title: Jessica Barboza - Facebook\newline
Content: Jessica Barboza ; Lives in \colorbox{black}{ABCDEFG} ; From São Paulo, Brazil ; In a relationship with \colorbox{black}{ABCDEFG}.\newline
Link: https://www.facebook.com/\colorbox{black}{ABCDEFG}\\\midrule

\multicolumn{1}{c}{\textbf{The claim is hard to verify because it is long and complex.}}\\\midrule

\textbf{Explanation}: The annotators did not agree on this item at all. The claim contains multiple parts that are correlated to each other. However, it is also hard to further break the claim down to smaller claims.\\

\textbf{Claim}: Chuck Norris's victory in the 1968 World Full-Contact Karate Championships solidified his reputation as one of the best martial artists in the world.\\\midrule 

\multicolumn{1}{c}{\textbf{Intermediate reasoning process is needed because the evidence might indirectly supports the claim.}}\\\midrule

\textbf{Explanation}: Two annotators chose the \supported~label and one chose \inconclusive. It is possible to interprete he ``safety objectives'' in search result 8 as it includes ``public health''. \\

\textbf{Claim}: The disposal of radioactive waste is aimed at ensuring public health.\\

\textbf{Evidence}:\newline
Search result 8\newline
Title: PART 61—LICENSING REQUIREMENTS FOR LAND DISPOSAL ...\newline
Content: (1) Disposal of radioactive waste in near-surface disposal facilities has the following safety objectives: protection of the general population from releases of ...\newline
Link: https://www.nrc.gov/reading-rm/doc-collections/cfr/part061/full-text.html
\\\bottomrule

\end{tabular}%
}
\caption{Example of the annotation items on which the annotators did not fully agree with each other. We mark out the private sensitive content.}
\label{tab:ex_of_disagree_cases}
\end{table*}

\subsection{Reason of disagreement in human verification}

There are 9 items in the human study in \sectionref{sec:claim.verification} on which the annotators did not reach a full agreement. After inspecting these items, we conclude 4 sources of disagreement, listed in  \tableref{tab:ex_of_disagree_cases} with examples. First, an annotator made a mistake (e.g., misread a name). Second, there is disagreement in the interpretation of the claim and evidence (e.g., \textit{can} in the claim vs.\ \textit{could} in the evidence or an ambiguous referent). Third, the claim is complex and long, hence, is hard to verify. Fourth, the evidence indirectly supports the claim which means intermediate reasoning process is needed. one annotator have overlooked the connection between the claim and the search result.

\subsection{Verifying/falsifying inconclusive claims is hard}

In \sectionref{sec:claim.verification}, we presented the distribution of verification labels in \tableref{tab:claim_veri_def_distribution}. As many as 42.2\% of the annotated items are labeled as \inconclusive~by our annotators. In order to understand whether the inconclusive cases can be verified/falsified by checking the full web page of the returned search results, we randomly picked 15 inconclusive cases and manually verified them. Results show that two claims are not specified enough to be verified, for example, \exampleref{ex.archaeologists}.

\begin{exe}
    \ex\label{ex.archaeologists}
    A group of archaeologists unearthed a cache of Roman weaponry near the ancient ruins of the Colosseum on a sweltering summer afternoon.
\end{exe}

Only one claim in \exampleref{ex.angular} can be verified the full web page. The search result snippets do not mention that Angular is maintained by Google but this is confirmed by a notice at the end of the web page.

\begin{exe}
    \ex\label{ex.angular} 
    Google Angular supports dependency injection.\\
    \href{https://angular.io/guide/architecture-services}{angular.io/guide/architecture-services}
\end{exe}

\noindent For the remaning 12 cases, we used Google search to find more evidence but only the claim in \exampleref{ex.battery} was weakly contradicted by a popular science article, which states the content in \exampleref{ex.battery_state}. If ``As a battery discharges'' is describing the status change of a battery from not discharging to discharging, the article implies that the chemical reactions become active but not slowing down. However, if the claim is understood as ``as a battery continues to discharge'', the article does not mention anything about chemical reactions slowing down gradually.

\begin{exe}
\ex
    \begin{xlist}
    \ex\label{ex.battery}
    As a battery discharges, the chemical reactions inside the battery slow down.
    
    \ex\label{ex.battery_state}
    When a battery is discharged, chemical reactions within the battery cells facilitate the movement of electrons from the negative terminal (anode) to the positive terminal (cathode) [...]. (\href{https://newsmartsafe.com/industry-news/battery-discharge}{link})
    \end{xlist}
\end{exe}

\section{Details of fine-tuning open-source models for claim extraction}
\label{appendix:ft_claim_extraction}

In this section, we specify the details of fine-tuning open-source models for the claim extraction task. 

\noindent \textbf{Data} We used two types of data for GPT-4 to extract claims, which are used to fine-tuned open-source models. The first type of data is the existing open-source data in \tableref{tab:seven_datasets_for_claim_extraction}. We sampled 100 data points from Scruples and 200 from the other datasets. The reason of sampling less data points from Scruples is that the majority of them is invariably subjective and yields the ``No verifiable claim.'' output. Hence, they are not very helpful in teaching an open-source model how to extract claims from factual texts. 

The second data type is our newly generated model responses. For this, we sampled 63 prompts from \fsbio~and 80 from the other datasets listed in \tableref{tab:seven_datasets_for_claim_extraction} (Usage = Dev). To generate the responses, we prompted the first 12 LLMs in \tableref{tab:simple_12_models_for_response_gen} with their default hyperparameters and the maximum token requirement was set to 1000. The LLMs are chosen in a way that we have both close- and open-source models as well as models in the same family with different sizes or versions. 

After collecting and generating all the model long-form responses, we decompose them into claims with GPT-4 using the prompts in \appendixref{appendix:our_claim_extraction_prompts}. The temperature was set to $0$.

We formed the fine-tuning data in the following way. As the input, we used the prompt (if there is one) and the response text with one sentence being marked with \texttt{<SOS>} and \texttt{<EOS>}. The output is the claims extracted from the marked sentence. 

To get the final set of fine-tuning data, we randomly removed $80\%$ of the data points whose marked sentence is shorter than 10 characters. These short marked sentences are usually the numbering of numbered lists. We also randomly dropped $50\%$ data whose output is ``No verifiable claim.'' In total, we got 99592 input-output pairs, among which 902 pairs have a short marked sentence, and 31819 pairs have ``No verifiable claim.'' as the output. We took $95\%$, $4\%$, and $1\%$ of the dataset as the training, validation, and test splits.




\noindent \textbf{Fine-tuning} We chose Llama3-8B-Instruct 
and Mistral-7B-Instruct-v0.2 as the base models (henceforth Llama3 and Mistral). Both were fine-tuned via Unsloth\footnote{\url{https://unsloth.ai/}} for two epochs using LoRA \citep{hu2022lora}. Checkpoints were saved at each epoch and tested on the test set. We used string-based metrics for evaluation and selected Mistral fine-tuned for two epochs as the best checkpoint.\footnote{The string-based metrics and the scores are the following: exact match $=0.4317$, Rouge1 $=0.8243$, Rouge2 $=0.7576$, RougeL $=0.8009$, and \textsc{chrF++} $=74.6686$.} We further evaluated this checkpoint manually.

\noindent \textbf{Manual quality comparison} To understand how good the performance of the fine-tuned model is compared to GPT-4, the first two authors did a pairwise comparison between the outputs from the two models on 300 test data points. After removing the data whose GPT-4 and Mistral outputs match exactly, there were 169 data points left for manual evaluation. Each annotator annotated the same data and were asked to choose which output was better or whether it was hard to choose between the two. 

\begin{figure}[t]
    \centering
    \includegraphics[scale=0.4]{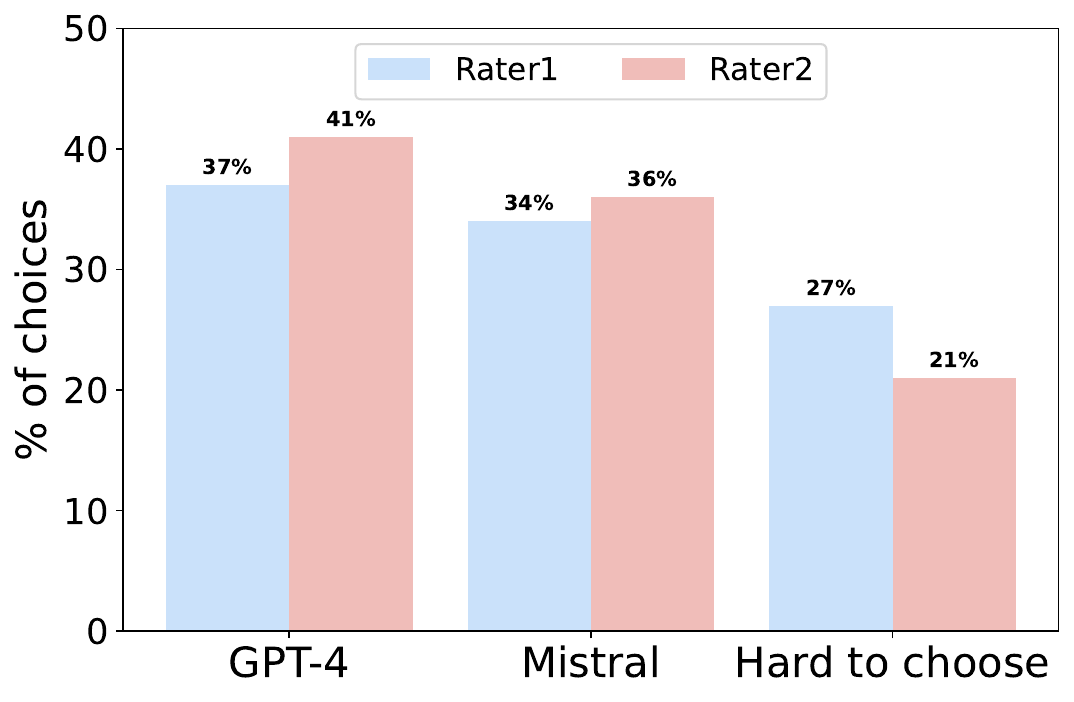}
    \caption{Results of comparing GPT-4 and fine-tuned Mistral-7B-Instruct-v0.2 on the claim extraction task. Numbers are in percentage. The Cohen's $\kappa$ between the two annotators is $0.4320$. Mistral achieves a competitive result compared to GPT-4.}
    \label{fig:gpt4_vs_mistral}
\end{figure}

\figureref{fig:gpt4_vs_mistral} shows the percentage of each annotator's choices. The two authors fully agreed in 106 data points, achieving a Cohen's $\kappa = 0.4320$ (moderate agreement, \citealp{landis1977measurement}).
Given that the quality of both models' outputs is close with GPT-4 being slightly better, such a moderate agreement is expected.

\noindent \textbf{Quality analysis} GPT-4 and the fine-tuned Mistral perform similarly. In many cases, their outputs are identical with certain phrases being relocated in the sentences. However, there are cases where Mistral lost to GPT-4 because it misses small words and makes the extracted claims less specific. Occasionally, Mistral puts multiple pieces of information into one claim while GPT-4 breaks the information down into multiple claims. Concrete examples can be seen in \tableref{tab:gpt4_mistral_quality_compare}. Besides these issues, in certain cases, Mistral does not refrain itself to the marked sentence in an input but also extracts claims that come after the span.

Overall, our fine-tuned Mistral model performs comparable to GPT-4. As an open-source model, it is also cost-efficient.

\begin{table*}[t]
\fontsize{8}{10}\selectfont
\resizebox{\textwidth}{!}{%
\begin{tabular}{@{}p{3.5cm}p{11cm}@{}}
\toprule
\textbf{Quality description}  & \textbf{Model outputs} (G = GPT-4, M = Mistral) \\ \midrule
Almost identical with certain phrases being relocated & \textbf{G}\hspace{1.2pt} Rubbing alcohol works by a different mechanism than antibiotics.\newline \textbf{M} Rubbing alcohol works by a  mechanism different than antibiotics. \\\midrule

Mistral misses small words, hence the extracted claims are less specific & \textbf{G}\hspace{1.2pt} Summarily laying off workers can have devastating impacts on individuals\newline \textbf{M} Laying off workers can have devastating impacts on individuals.\\ \midrule

Mistral puts multiple pieces of information into one claim while GPT-4 breaks the information down into multiple claims & \textbf{G}\hspace{1.2pt} (1) The regimental flags used during Napoleon’s invasion of Russia by Bavaria were similar to their design. (2) The regimental flags for Bavaria during Napoleon’s invasion of Russia had a green wreath added around the eagle. \newline \textbf{M} Bavarian regimental flags used during Napoleon’s invasion of Russia in 1812 featured a green wreath around the eagle.\\\bottomrule

\end{tabular}%
}
\caption{Quality analysis of the outputs generated by prompted GPT-4 and fine-tuned Mistral-7B-Instruct-v0.2. The table lists three common observations made in the pairwise comparison task.}
\label{tab:gpt4_mistral_quality_compare}
\end{table*}

\section{Details of fine-tuning open-source models for claim verification}\label{appendix:ft_claim_verification}

In this section, we offer details of fine-tuning the open-source models for the claim verification task, in addition to \sectionref{subsec:fine-tuned_models}. 

In order to select the best model for generating fine-tune data, we tested Mixtral-8$\times$22-Instruct-v0.1, Claude-3-Opus, GPT-4, and GPT-4o with the few-shot prompt in \tableref{tab:claim_verification_prompt} on the 320 human annotated verification data in \sectionref{sec:claim.verification}. We tried two types of verification task, one with \supported~and \unsupported~as the labels and one with \supported, \contradicted, and \inconclusive~as the labels. For a fair comparison between the model performance on the binary and ternary task, we convert the \contradicted~and \inconclusive~labels in the ternary task to \unsupported. We then calculated the F1 scores on all 320 items as well as on \supported~and \unsupported~items separately. The results are in \tableref{tab:prompted_model_on_320}. Overall, GPT-4o with ternary labels has the most balanced performance on the \supported~and \unsupported~items. Hence, we generated data from GPT-4o for fine-tuning open-source models and converted the ternary labels to binary ones.

\begin{table}[t]
\fontsize{8}{10}\selectfont
\centering
\resizebox{0.95\columnwidth}{!}{%
\begin{tabular}{@{}p{1cm}p{0.9cm}p{1.25cm}p{0.7cm}p{0.7cm}@{}}
\toprule
\textbf{Model}  & \textbf{Label \#} & \textbf{Overall F1} & \textbf{S. F1} & \textbf{U. F1} \\ \midrule
Mixtral & 2 & 0.817 & 0.817 & 0.686 \\
Mixtral & 3 & 0.807 & 0.807 & 0.680 \\
Claude 3 & 2 & 0.839 & 0.839 & \textbf{0.758} \\
Claude 3 & 3 & 0.826 & 0.826 & 0.681 \\
GPT-4    & 2 & 0.829 & 0.829 & 0.696  \\
GPT-4    & 3 & 0.812 & 0.812 & 0.639  \\
GPT-4o  & 2 & 0.813 & 0.813 & 0.649 \\
GPT-4o  & 3 & \textbf{0.841} & \textbf{0.841} & 0.731 \\\bottomrule
\end{tabular}%
}
\caption{The results of testing four prompted LLM on the claim verification task. S.\ = \supported~and U.\ = \unsupported. Mixtral stand for Mixtral-8x22B-Instruct-v0.1. Claude 3 stands for Claude 3 Opus. GPT-4o achieves the highest overall F1 and F1 on the items with \supported~as the label. Although its \unsupported~F1 is not the highest, it is not far below the highest, which is Claude 3 with 0.758.}
\label{tab:prompted_model_on_320}
\end{table}

\noindent \textbf{Data} We sampled 10 prompts from the datasets listed in \tableref{tab:seven_datasets_for_claim_extraction} (Usage = Dev) and prompted the LLMs in \tableref{tab:simple_12_models_for_response_gen} with their default hyperparameters and the maximum token requirement was set to 1024. The model responses are decomposed into claims with GPT-4 using the prompts in \appendixref{appendix:our_claim_extraction_prompts}. The temperature was set to $0$. Serper is then used to retrieve search results as described in \sectionref{subsec:define_evidence_retrieval}. As the prompt, we use the binary prompt template in \tableref{tab:claim_verification_prompt} without the few-shot examples. From the generated data, we randomly sampled 13403 data points, among which 9996 has the \supported~label and 3407 has the \unsupported~label. We split the dataset into 85\%, 3\%, and 12\% as the training, validation, and test splits. 

\noindent \textbf{Fine-tuning} Similar to the fine-tuned claim extractor, we fine-tuned Llama3-8B-Instruct and Mistral-7B-Instruct-v0.2 via Unsloth for 5 epochs. Because the number of \supported~and \unsupported~items in the training dataset are imbalanced, we triplicated the \unsupported~data points. Checkpoints were saved at each epoch and tested on both the test dataset and the 320 human annotated set. The Llama3-8B-Instruct model fine-tuned for one epoch achieves the most balanced performance on \supported~and \unsupported~data points on the test and human data. Hence, we use this checkpoint for further experiments.

\begin{table*}[ht]
\fontsize{10}{12}\selectfont
\centering 
\resizebox{\textwidth}{!}{%
\begin{tabular}{@{}p{20cm}@{}}
\toprule
\multicolumn{1}{c}{\textbf{\scalebox{1.1}{Prompt for verifying claims with three labels}}}
\\\midrule

You need to judge whether a claim is supported or contradicted by Google search results, or whether there is no enough information to make the judgement (i.e., inconclusive). When doing the task, take into consideration whether the link of the search result is of a trustworthy source. Mark your answer with \#\#\# signs. 
\newline

Below are the definitions of the three categories:

Supported: A claim is supported by the search results if everything in the claim is supported and nothing is contradicted by the search results. There can be some search results that are not fully related to the claim.

Contradicted: A claim is contradicted by the search results if something in the claim is contradicted by some search results. There should be no search result that supports the same part.

Inconclusive: A claim is inconclusive based on the search results if:

- a part of a claim cannot be verified by the search results,

- a part of a claim is supported and contradicted by different pieces of evidence,

- the entity/person mentioned in the claim has no clear referent (e.g., "the approach", "Emily", "a book").
\newline

Here are some examples:
\newline

Claim: Vikings used their longships to transport livestock.

Search result 1

Title: How did the Vikings transport animals on their ships? - Quora

Content: The Vikings transported horses overseas in boats very similar to Viking longships, but with flat flooring built within the hulls, which allowed ...

Link: https://www.quora.com/How-did-the-Vikings-transport-animals-on-their-ships

Your decision: \#\#\#Contradicted.\#\#\#

<Other search results omitted for the sake of space>
\newline

<nine such examples>
\newline

Your task:

Claim: \{claim to be verified\}

\{search results\}

Your decision:\\\midrule

\multicolumn{1}{c}{\textbf{\scalebox{1.1}{Prompt for verifying claims with two labels}}}
\\\midrule

Everything being the same but the definitions of the labels are changed as below. The decisions in the few-shot exmaples are converted to \supported~and \unsupported~accordingly (i.e., \contradicted~and \inconclusive~become \unsupported).
\newline

Supported: A claim is supported by the search results if everything in the claim is supported and nothing is contradicted by the search results. There can be some search results that are not fully related to the claim.

Unsupported: If a claim is not supported by the search results, mark it as unsupported.\\\bottomrule

\end{tabular}%
}
\caption{Claim verification prompt for Mixtral-8$\times$22B-Instruct-v0.1, GPT-4, and GPT-4o. For Claude 3 Opus, the order of the claim, search results, and the decision is rearranged. Otherwise, the model does not always output a decision marked by \#\#\#. The rearranged order is search results, the claim, a short task description, and the decision. The short task description is ``Task: Given the search results above, is the claim \{supported, contradicted, or inconclusive\}/\{supported or unsupported\}? Mark your decision with \#\#\# signs.'' The set of labels in the curly brackets depends on whether the verificaton task is binary or trinary.}


\label{tab:claim_verification_prompt}
\end{table*}

\section{Details of data domains and studied LLMs}
\label{appendix:data_and_llms}

This section gives the details of the datasets and LLMs in \sectionref{subsec:data_and_lms}. \tableref{tab:seven_datasets_for_claim_extraction} lists the datasets that are used for developing, test, and benchmark models on \name. \tableref{tab:new_book_dataset} further expands the name and details of the FreshBooks dataset. For developing \name, we used the model generations from the first 12 models in \tableref{tab:simple_12_models_for_response_gen}. For benchmarking models on \name, we used all 16 models in \tableref{tab:simple_12_models_for_response_gen}.

\begin{table*}[!ht]
\fontsize{6}{8}\selectfont
\resizebox{\textwidth}{!}{%
\begin{tabular}{@{}p{6cm}p{4.5cm}p{1.5cm}@{}}
\toprule
\textbf{Book name}  & \textbf{Author/editor/translator}   & \textbf{Publication date} \\\midrule
Blunt Instruments: Recognizing Racist Cultural Infrastructure in Memorials, Museums, and Patriotic Practices & Kristin Ann Hass & January, 2023 \\\midrule
Every Living Thing: The Great and Deadly Race to Know All Life & ason Roberts  & April, 2024 \\\midrule
It's OK to Be Angry About Capitalism  & Bernie Sanders, John Nichols & 2024 \\\midrule
Out of the Darkness: The Germans, 1942-2022  & Frank Trentmann  & February, 2024 \\\midrule
Takeover: Hitler's Final Rise to Power & Timothy W. Ryback & March, 2024 \\\midrule
The Exhausted of the Earth: Politics in a Burning World & Ajay Singh Chaudhary & February, 2023 \\\midrule
The Making of a Leader: The Formative Years of George C. Marshall  & Josiah Bunting III  & March, 2024 \\\midrule
They Were Here Before Us: Stories from the First Million Years  & Eyal Halfon, Ran Barkai & March, 2024 \\\midrule
The Green Power of Socialism: Wood, Forest, and the Making of Soviet Industrially Embedded Ecology & Elena Kochetkova  & February, 2024  \\\midrule
A Brief History of Feminism  & Patu, Antje Schrupp, Sophie Lewis & April, 2024 \\\midrule
Handbook of Formal Analysis and Verification in Cryptography & Sedat Akleylek, Besik Dundua  & September, 2023 \\\midrule
Handbook on Renewable Energy and Green Technology  & S. Pugalendhi, J. Gitanjali, R. Shalini, P. Subramanian & February, 2024 \\\midrule
The Handbook of Sex Differences Volume I Basic Biology  & Lee Ellis, Craig T. Palmer, Rosemary Hopcroft, Anthony W. Hoskin & September, 2023 \\\midrule
The Oxford Handbook of Thomas More's Utopia & Cathy Shrank, Phil Withington & February, 2024 \\\midrule
The Routledge Handbook of Commodification & Elodie Bertrand, Vida Panitch  & December, 2023 \\\midrule
The Routledge Handbook of Green Finance & Othmar M. Lehner, Theresia Harrer, Hanna Silvola, Olaf Weber & November, 2023 \\\midrule
The Routledge Handbook of Language and Religion  & Stephen Pihlaja, Helen Ringrow & December, 2023 \\\midrule
The Routledge Handbook of Language and Youth Culture   & Bente A. Svendsen, Rickard Jonsson & December, 2023 \\\midrule
Clinical Handbook of Nephrology & Robert S. Brown MD & August, 2023 \\\midrule
Handbook of Face Recognition: The Deep Neural Network Approach  & Stan Z. Li, Anil K. Jain, Jiankang Deng & 2024 \\\bottomrule
\end{tabular}%
}
\caption{Twenty newly published non-fictional books. We took ten paragraphs from each book and used them to prompt LLMs to generate continuations. The selected paragraphs are all located at the beginning of a chapter/section.}
\label{tab:new_book_dataset}
\end{table*}

\begin{table}[t]
\fontsize{6}{8}\selectfont
\resizebox{0.98\columnwidth}{!}{%
\begin{tabular}{@{}p{2cm}p{0.6cm}p{2.2cm}@{}}
\toprule
\textbf{Model Name} & \textbf{Release}  & \textbf{Reference} \\ \midrule

GPT-3.5-turbo & 2023.06 & \multirow{3}{*}{\citet{DBLP:journals/corr/abs-2303-08774}}\\

GPT-3.5-turbo & 2023.11 & \\

GPT-4 & 2024.01 &  \\\midrule

Claude-3-Haiku & 2024.03 & \multirow{3}{*}{\citet{Anthropic2023Claude}} \\

Claude-3-Sonnet & 2024.02  & \\

Claude-3-Opus  & 2024.02 & \\\midrule

Mist-7B-Inst-v0.1  &  2023.09 & \multirow{4}{*}{\citet{Jiang2023Mistral7, Jiang2024MixtralOE}} \\

Mist-7B-Inst-v0.2  & 2023.12  & \\

Mixt-8$\times$7B-Inst-v0.1 &  2023.12  &  \\

Mixt-8$\times$22B-Inst-v0.1 & 2024.04  & \\\midrule

OLMo-7B-Inst & 2024.01 & \citet{groeneveld2024olmo} \\\midrule

DBRX Inst (132B) &  2024.03 & \citet{DBRXblog}\\\midrule

Qwen1.5-1.8B-Chat & 2023.11 & \citet{bai2023qwen} \\

Gemma-2B-it& 2024.04 & \citet{gemmateam2024gemma}\\

Vicuna-7B-v1.5 & 2023.12 & \citet{zheng2023judging}\\

GPT-4o & 2024.05 & \citet{openai-gpt4o} \\\bottomrule
\end{tabular}%
}
\caption{Sixteen models that are tested in the current work. Inst, Mist, and Mixt stands for instruction, Mistral, and Mixtral. The numbers of total parameters of each model are given in brackets if provided by the model providers and not in the model names. The first 12 models are also used to generate model responses for fine-tuning claim extraction and verification models.}
\label{tab:simple_12_models_for_response_gen}
\end{table}

\section{\texttt{Unsupported} cases in \name\ outputs}\label{appendix:veriscore_inconclusive}

This appendix section provides more examples of the \unsupported~claims in \tableref{tab:qualitative_examples} in complementary to \sectionref{subsubsec:search_result_insufficient}.

\begin{table*}[ht]
\fontsize{9}{11}\selectfont
\centering 
\resizebox{\textwidth}{!}{%
\begin{tabular}{@{}p{16cm}@{}}
\toprule

\multicolumn{1}{c}{\textbf{\scalebox{1.1}{Claim that is unsupported because it uses a different term as its search results}}}
\\\midrule

\textbf{Claim}: In long trains, additional locomotives can be placed along the train to help distribute the pulling force more evenly. (ELI5)

\textbf{Search result}

Title: Nuts \& Bolts: Why is there an engine in the middle of that train?

Content: By placing DPUs\textsuperscript{*} throughout the train rather than just at the rear—thus distributing power more evenly—railroads were able to enhance a train's ...

Link: \href{https://gorail.org/infrastructure/nuts-bolts-why-is-there-an-engine-in-the-middle-of-that-train}{https://gorail.org/infrastructure/nuts-bolts-why-is-there-an-engine-in-the-middle-of-that-train}

\textbf{*} DPU stands for Distributed Power Unit, a locomotive set.\\\midrule

\multicolumn{1}{c}{\textbf{\scalebox{1.1}{Unsupported claims that have reasonable content}}}\\\midrule

- Spreading the load across multiple axles reduces the stress on individual components in trains. (ELI5)

- Detailed interviews with patients and their contacts help to establish timelines. (ELI5)\\\midrule

\multicolumn{1}{c}{\textbf{\scalebox{1.1}{Unsupported claims that are too vague to be verified}}}
\\\midrule

- Missiles travel through the Earth's atmosphere for most or all of their flight. [Not clear which type of missiles] (ELI5) 

- The forests of Siberia and the Far East were crucial for meeting the demand for wood and wood products for export. [Not clear whose demand it is] (FreshBooks)\\\midrule

\multicolumn{1}{c}{\textbf{\scalebox{1.1}{Unsupported claims that are too broad to be verified}}}
\\\midrule

- Marshall's leadership and strategic acumen ensured the maneuver was carried out flawlessly during a field maneuver in the Philippines. (FreshBooks)

- Germany is maintaining its competitive edge in a rapidly changing global landscape. (FreshBooks)

- The initiatives of the General German Women's Union helped to lay the groundwork for future advancements in women's rights in Germany. (FreshBooks)\\\midrule

\multicolumn{1}{c}{\textbf{\scalebox{1.1}{General unsupported claims---no supporting evidence}}}\\\midrule

- The movement of the sun or stars could be compared with the rate of flow in water clocks. (ELI5)

- Driving from the middle of a car complicates interactions with road design elements.  (ELI5)

- The patronage of William Warham reflects the broader trend of Renaissance humanism gaining foothold in England. (FreshBooks)\\\bottomrule

\end{tabular}%
}
\caption{Types and reasons of claims being \unsupported.}
\label{tab:qualitative_examples}
\end{table*}

\section{\name\ on WritingPrompts and FreshQA}\label{appendix:WP_and_FreshQA}

In \sectionref{sec:veriscore_16_models_6_domains}, we presented \name\ of 16 models on 6 domains of long-form model generation. In this section, we focus solely on model responses in the FreshQA and WritingPrompts datasets. As shown in \tableref{tab:grand_table}, both domains yield very few verifiable facts. The median number for verifiable claims ($K$) in FreshQA is four because the questions in general do not require long-form answers. WritingPrompts requires long-form generations but they are conditioned on fictional premises. Hence, the generations contain very few verifiable claims, resulting in $K = 1$.


For the generations in WritingPromtps, the results in \tableref{tab:grand_table} show that the \name\ of the models is very low, matching the expectation that there are few supported verifiable claims in a creative writing task. We examine the extracted claims from two models---Gemma-2b-it and GPT-4o. It turns out that the majority of the claims (82.61\% for Gemma and 71.15\% for GPT-4o) has the potential to be verified/falsified. The results of the models on WritingPrompts prove that the \name\ pipeline works effectively on fictional content although it occasionally extracts unverifiable claims.

For the generations in FreshQA, the results show that the Claude 3 models perform the best. However, upon careful examination of the outputs of Claude 3 Haiku and GPT-4o, we notice that GPT-4o has a higher percentage of supported claim among all the extracted claims (74.70\%) compared to Claude 3 Haiku (72.71\%). The fact that GPT-4o generates shorter outputs than Claude 3 Haiku contributes to the lower final \name\ of GPT-4o. On average, Claude 3 Haiku has 5.5 claims per response, higher than $K = 4$, but GPT-4o has only 3.39. After reading the claims extracted from both models in the FreshQA domain, we notice that GPT-4o tends to generate short and to-the-point answers while Claude 3 Haiku tends to generate longer answers, offering more auxiliary information. Within the Claude 3 model family, Claude 3 Haiku generates longer outputs than the other two and has more responses with 4 or more supported claims, resulting in a higher \name.

The results of FreshQA shows that, for a fair comparison, when calculating \name, the length of model responses should be taken into account. This can be done by forcing all the models to generate a similar length of outputs. However, this will not result in a setting of how end-users would use language models. Forcing models to generate longer responses than necessary can also elicit more infactual content, as noticed by \citet{factscore} that later content in model responses tends to be less factual.

\section{Detailed results of models' \name}\label{appendix:grand_table}

In this appendix section, we provide the breakdown of \name\ of the 16 models on all 8 domains in \tableref{tab:grand_table}. 

\begin{table*}[t]
\centering

\resizebox{1\textwidth}{!}{%
\begin{tabular}{@{}p{3.45cm}p{0.55cm}p{0.55cm}p{0.55cm}p{0.55cm}|p{0.55cm}p{0.55cm}p{0.55cm}p{0.55cm}|p{0.55cm}p{0.55cm}p{0.55cm}p{0.55cm}|p{0.55cm}p{0.55cm}p{0.55cm}p{0.55cm}|p{0.55cm}p{0.55cm}p{0.55cm}p{0.55cm}|p{0.55cm}p{0.55cm}p{0.55cm}p{0.55cm}|p{0.55cm}p{0.55cm}p{0.55cm}p{0.55cm}|p{0.55cm}p{0.55cm}p{0.55cm}p{0.58cm}@{}}
\toprule
\multirow{2}{*}{\textbf{Model}}  & \multicolumn{4}{c}{\textbf{LongFact (32)}} & \multicolumn{4}{c}{\textbf{\fsbio~(26)}} & \multicolumn{4}{c}{\textbf{ELI5 (21)}} & \multicolumn{4}{c}{\textbf{AskHist (21)}} & \multicolumn{4}{c}{\textbf{FreshBooks (24)}} & \multicolumn{4}{c}{\textbf{ShareGPT (11)}} & \multicolumn{4}{c}{\textbf{FreshQA (4)}} & \multicolumn{4}{c}{\textbf{WP (1)}} \\

 & L & P & R & F & L & P & R & F & L & P & R & F & L & P & R & F & L & P & R & F & L & P & R & F & L & P & R & F & L & P & R & F \\ \midrule

Gemma-2b-it & 20.4 & 67.2 & 61.4 & 60.7 & 11.8 & 4.3 & 5.1 & 4.6 & 8.7 & 38.6 & 27.2 & 28.8 & 6.6 & 28.2 & 17.2 & 17.8 & 5.0 & 43.1 & 20.0 & 25.1 & 15.4 & 28.8 & 32.7 & 27.6 & 1.4 & 8.7 & 4.0 & 4.8 & 23.8 & 3.2 & 6.1 & 3.7\\
Qwen1.5-1.8B-Chat & 22.1 & 64.4 & 78.2 & 70.3 & 16.2 & 11.4 & 18.5 & 14.1 & \textbf{21.2} & 49.1 & 74.0 & 57.9 & 16.6 & 37.7 & 59.2 & 45.2 & 19.4 & 45.1 & 64.8 & 52.6 & 24.1 & 42.7 & \textbf{64.7} & 49.2 & \textbf{7.1} & 35.9 & 68.5 & 43.9 & 29.5 & 4.1 & 10.0 & 5.5\\
Vicuna-7b-v1.5 & 12.1 & 79.3 & 57.0 & 63.4 & 8.8 & 26.3 & 21.9 & 23.0 & 8.3 & 59.8 & 47.5 & 51.3 & 8.7 & 43.3 & 39.7 & 39.7 & 7.1 & 50.2 & 35.0 & 39.0 & 15.1 & 46.6 & 45.8 & 43.6 & 2.6 & 42.5 & 29.5 & 30.9 & 20.1 & 2.0 & 2.0 & 2.0\\
OLMo-7B-Inst & 21.2 & 72.3 & 77.3 & 73.4 & \textbf{16.8} & 17.4 & 22.7 & 19.4 & 18.3 & 55.0 & 65.2 & 58.8 & 16.7 & 41.5 & 50.5 & 43.2 & \textbf{24.1} & 49.9 & 60.6 & 53.7 & 25.5 & 48.0 & 60.5 & 49.4 & 3.6 & 46.1 & 56.5 & 49.2 & 31.2 & 2.0 & 2.0 & 2.0\\
DBRX-Inst & 15.6 & 85.0 & 72.3 & 75.9 & 13.2 & 45.4 & 48.6 & 46.5 & 11.5 & 69.0 & 59.3 & 61.9 & 13.6 & 49.7 & 52.2 & 49.5 & 13.1 & 58.9 & 62.6 & 60.2 & 18.0 & 49.0 & 54.0 & 48.9 & 4.3 & 66.1 & 69.5 & 64.9 & 26.4 & \textbf{12.1} & \textbf{18.0} & \textbf{13.3}\\
Mist-7B-Inst-v0.1 & 10.3 & 76.5 & 50.6 & 57.6 & 10.3 & 22.1 & 19.6 & 20.3 & 7.0 & 59.2 & 35.2 & 42.2 & 8.0 & 45.1 & 33.3 & 36.5 & 7.1 & 53.3 & 34.7 & 39.8 & 16.3 & 46.6 & 43.3 & 41.2 & 1.8 & 46.8 & 24.0 & 29.0 & 21.2 & 6.4 & 10.0 & 7.3\\
Mist-7B-Inst-v0.2 & 16.0 & 83.2 & 68.4 & 72.0 & 11.9 & 29.6 & 31.2 & 30.0 & 11.3 & 63.9 & 58.7 & 58.8 & 14.5 & 43.9 & 45.4 & 41.2 & 10.7 & 54.6 & 52.4 & 52.4 & 19.1 & \textbf{54.0} & 61.1 & \textbf{54.8} & 4.1 & 57.9 & 55.5 & 54.2 & 29.8 & 2.6 & 6.0 & 3.4\\
Mixt-8x7B-Inst-v0.1 & 17.7 & 84.4 & 75.1 & 77.3 & 11.1 & 42.2 & 44.0 & 42.5 & 11.6 & 68.0 & 60.7 & 61.9 & 13.8 & 50.9 & 54.3 & 50.7 & 10.4 & 59.8 & 56.8 & 57.4 & 17.7 & 53.0 & 56.4 & 51.5 & 3.4 & 59.8 & 52.5 & 53.1 & 30.7 & 5.3 & 8.0 & 5.8\\
Mixt-8x22B-Inst-v0.1 & 17.6 & 86.7 & 76.2 & 78.0 & 12.6 & 47.4 & 49.2 & 47.6 & 12.4 & \textbf{69.1} & 64.6 & 64.9 & 13.3 & 52.5 & 54.0 & 51.1 & 12.4 & 60.3 & 58.3 & 58.0 & 19.8 & 50.4 & 57.3 & 51.4 & 3.3 & 65.6 & 58.0 & 59.6 & 30.4 & 5.4 & 8.0 & 6.1\\
Claude-3-haiku & 17.4 & 85.9 & 76.4 & 79.4 & 8.0 & 42.9 & 35.4 & 37.1 & 14.8 & 60.5 & 60.5 & 58.7 & 13.8 & 42.8 & 46.3 & 43.5 & 9.0 & 53.9 & 47.3 & 49.5 & 18.0 & 44.8 & 51.1 & 44.7 & 5.5 & \textbf{75.3} & \textbf{89.0} & \textbf{77.9} & 24.6 & 4.0 & 4.0 & 4.0\\
Claude-3-sonnet & 19.4 & 85.8 & 78.4 & 80.7 & 7.9 & 44.2 & 36.0 & 37.6 & 13.7 & 58.9 & 56.3 & 56.2 & 13.1 & 39.6 & 43.6 & 40.7 & 10.6 & 59.6 & 60.1 & 59.3 & 16.2 & 50.4 & 58.5 & 51.7 & 4.9 & 71.3 & 86.5 & 76.2 & 25.6 & 2.1 & 4.0 & 2.2\\
Claude-3-opus & 21.4 & 88.3 & 81.9 & 83.6 & 10.5 & 51.8 & 54.4 & 52.7 & 14.3 & 66.6 & 63.3 & 63.4 & 14.8 & 49.2 & 52.9 & 49.8 & 12.2 & 63.3
& 70.5 & 66.4 & 19.0 & 50.9 & 56.5 & 51.6 & 5.4 & 72.2 & 81.0 & 72.2 & 24.7 & 1.3 & 4.0 & 1.9\\
GPT-3.5-turbo-0613 & 16.9 & 87.7 & 75.1 & 77.6 & 14.5 & 43.9 & 49.2 & 45.9 & 13.1 & 67.8 & 63.2 & 62.9 & 12.9 & 52.9 & 53.6 & 51.8 & 11.7 &
50.3 & 51.3 & 49.0 & 19.1 & 51.1 & 52.5 & 48.6 & 1.7 & 68.4 & 38.5 & 44.9 & 37.2 & 4.3 & 6.0 & 4.9\\
GPT-3.5-turbo-1106 & 10.2 & \textbf{90.8} & 56.3 & 64.7 & 5.3 & \textbf{54.1} & 30.6 & 38.1 & 6.4 & 61.9 & 35.3 & 42.8 & 7.8 & 51.8 & 37.0 & 40.8 & 4.3 & 51.5 & 24.8 & 32.5 & 13.3 & 47.9 & 43.1 & 42.1 & 1.5 & 65.7 & 36.5 & 44.0 & 24.2 & 4.0 & 8.3 & 5.0\\
GPT-4-0125-preview & 20.6 & 84.3 & 89.2 & 85.9 & 13.0 & 52.2 & \textbf{63.8} & 56.4 & 20.1 & 63.7 & \textbf{83.0} & 70.7 & 18.8 & 47.8 & 72.6 & 56.6 & 12.8 & 64.6 & \textbf{77.0} & 69.7 & 23.6 & 51.5 & 60.2 & 53.5 & 2.9 & 67.8 & 59.5 & 60.3 & 33.3 & 3.1 & 6.0 & 3.7\\
GPT-4o & \textbf{25.8} & 85.4 & \textbf{89.8} & \textbf{86.7} & 13.1 & 53.5 & 61.5 & \textbf{56.7} & 20.4 & 67.1 & 79.8 & \textbf{71.7} & \textbf{22.6} & \textbf{54.2} & \textbf{77.9} & \textbf{61.4} & 11.6 & \textbf{68.8} & 75.2 & \textbf{70.9} & \textbf{28.2} & 49.3 & 56.5 & 51.5 & 2.6 & 67.9 & 55.0 & 58.2 & \textbf{48.6} & 5.3 & 8.0 & 6.2\\\bottomrule
\end{tabular}%
}
\caption{Details of \name\ of 16 models on all 8 domains. The maximum values for each metric in every category are highlighted in bold. L = average sentence count per response; P = average response precision; R = average response recall; F = \name. AskHist = AskHistorians; WP = WritingPrompts.}
\label{tab:grand_table}
\end{table*}

\end{document}